\documentclass[10pt,twocolumn,letterpaper]{article}

\usepackage[pagenumbers]{cvpr} 

\usepackage{cuted} 

\usepackage{currfile} 

\usepackage{microtype}

\renewcommand{\paragraph}[1]{\vspace{.5em}\noindent\textbf{#1}}

\DeclareMathOperator*{\search}{search}

\usepackage{soul}
\setuldepth{foobar}

\usepackage{nth} 
\usepackage{enumitem} 
\usepackage{float} 
\usepackage{siunitx}          
\usepackage{pifont}        

\newcommand{\cmark}{\ding{51}}   
\newcommand{\xmark}{\ding{55}}   

\usepackage{xspace}
\newcommand{\bestofn}{Best-of-$N$\xspace}
\usepackage{chapterbib}

\usepackage{adjustbox}
\usepackage{pdflscape}
\usepackage{multirow}
\usepackage[table]{xcolor}

\usepackage{tabularx}
\newcommand{\model}[1]{\textnormal{#1}}

\usepackage[most]{tcolorbox}
\newtcolorbox{promptbox}{ colback=gray!5, colframe=gray!45, boxrule=0.5pt, arc=2pt, left=6pt, right=6pt, top=5pt, bottom=5pt, fontupper=\small, breakable}

\usepackage{CJKutf8}

\definecolor{cvprblue}{rgb}{0.21,0.49,0.74}
\usepackage[pagebackref,breaklinks,colorlinks,allcolors=cvprblue]{hyperref}

\def\confName{CVPR}
\def\confYear{2027}

\title{Is model self-assessment enough for diverse LVLM ensembles?}

\author{Michael Smith \quad Frank P. Ferrie\\
Centre for Intelligent Machines, McGill University\\
{\tt\small michael.smith6@mail.mcgill.ca \quad  frank.ferrie@mcgill.ca}
}

\begin{document}

\maketitle
\begin{abstract}
Ensembles of diverse Large Vision Language Models have demonstrated performance beyond that of single models across multiple datasets and tasks. However, the necessary aggregation process usually relies on training, parameter fitting, or complex task-specific pipelines. Furthermore, evaluation is limited, making a fair comparison difficult. We create a unified pipeline where we perform candidate response selection: we build an ensemble by selecting the best answer from a set provided by different models. We comprehensively evaluate across 34 different means of confidence estimation, 24 ensemble configurations, 20 datasets, and two prompt specifications. We show that simple training and parameter-free approaches can be sufficient, even for ensembles of different architectures. These simple self-assessment approaches rely solely on each constituent model's confidence, without explicit calibration, yet are able to perform better than said constituent models. They are also competitive with more involved alternatives on standard metrics such as accuracy, including on open-ended problems. However, we also discover that dataset, ensemble composition, and prompt wording are key confounding variables insufficiently explored by prior work that significantly affect ensemble performance, such that the ensemble can perform worse than its best constituent model. These must be optimized for to get the most out of ensembles, at the cost of limiting their generalizability. We recommend that practitioners consider training-free ensembles only if deployment conditions can be precisely characterized, with \bestofn configurations of strong models being a good starting point.
\end{abstract}    
\section{Introduction}
\label{sec:intro}

\begin{figure*}[ht]
    \centering
    \includegraphics[width=0.67\linewidth]{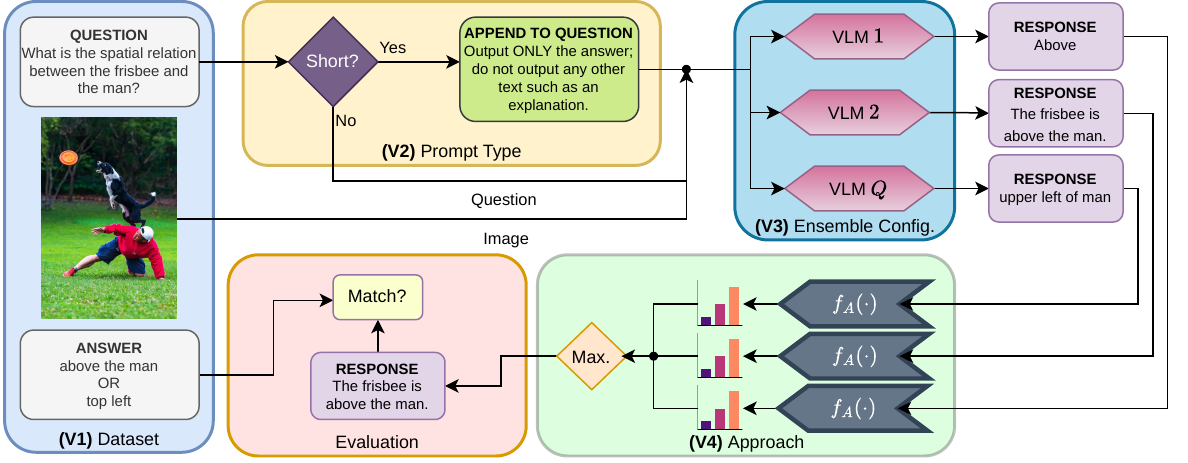}
    \caption{An overview of the testbed, showing the variables \labelcref{v:dataset,v:prompt,v:ensemble,v:approach} that can affect how one example question from the MMVet dataset is processed through to evaluation. See \cref{subsec:testbed_design} for details and \cref{subsec:variables} for a description of the variables.}
    \label{fig:meth:testbed}
\end{figure*}

The long-standing concept of ensembles has the goal of pushing performance beyond what can be achieved with a single model \cite{zhou_ensemble_2025}. More recently, Large Language Models (LLMs) and Large Vision Language Models (LVLMs) have exploded in popularity, including open models such as the Qwen 3 series \cite{team_qwen35_2026}. Focusing on LVLMs, it has been shown that across a selection of datasets, one model may lead on many, but not \textit{all}, datasets \cite{duan_vlmevalkit_2024,he_mordal_2026}. This has led to much work on obtaining the best performance possible across a range of scenarios, without the effort of training or fine-tuning a model for each one. Examples include model selection \cite{he_mordal_2026}, routing \cite{guha_smoothie_2024} and, of course, ensembles.

Ensembles have been used for decades \cite{zhou_ensemble_2025} and continue to be used with Large Language Models \cite{jiang_llm-blender_2023} and Large Vision Language Models \cite{yu2026scoop,tekin_vision_2026} but recent work suffers from two primary limitations. First, the aggregation step in the ensemble is usually addressed in a way that imposes substantial requirements. Examples include training or parameter fitting \cite{pandey_refine_2025, tekin_vision_2026}, assuming that the task is compatible with the clustering of answers \cite{yu2026scoop,yu_cusp_2026,wang_seeing_2026,zhang_consensus_2026,tong_diversity_2026} or adding substantial task-specific complexity \eg VQA \cite{wang_seeing_2026} or OCR \cite{zhang_consensus_2026}. One of the strengths of LVLMs is the ability for one model to be used for many different tasks, and with ensembles specifically we want to maximize performance on all of them. These requirements undercut this strength by limiting the ensemble to specific problem types or datasets. Second, evaluation is usually constrained by a limited selection of benchmarks or baseline comparisons, high variation in baseline performance, or some combination thereof \cite{pandey_refine_2025,alazraki_how_2023, tekin_vision_2026, yu2026scoop, fu_deep_2026, zollicoffer_mtre_2025,park_vauq_2026, yao_reading_2025,wang_vlm-uqbench_2026}. LVLM ensembles are not well studied in general, with much more attention devoted to Large Language Models \cite{xia_survey_2025,chen_harnessing_2026,boudiaf_uncertainty-aware_2026} or to approaches that rely on a single model \cite{yao_reading_2025, zollicoffer_mtre_2025}. Overall, it is unclear from the literature if useful (performing better than a single model) LVLM ensembles can be constructed \textit{without} imposing burdens such as training or a limited problem set.

In this paper, we develop a testbed that allows us to study the performance of LVLM ensembles, with an eye to answering several research questions surrounding them. We ensure many variables, including prompt changes, ensemble configuration and dataset are well explored. We also maintain the general applicability of each ensemble by formulating the ensemble task as one of \textbf{candidate response selection} \ie selecting the best answer to a given question from a set, with each answer generated by a different model. We ensure a fair comparison by examining only training-free approaches compatible with open-ended questions, specifically those that rely on a model's self-assessment. We find that simple approaches such as those that use a model's token-level confidence measurements allow for an ensemble to perform better than the best single model. Furthermore, these approaches can be competitive with those that are much more complex or rely on training data. However, variables such as ensemble composition or dataset play a very important role in how an ensemble performs, and must be optimized for. Thus, there are limits as to how broadly applicable a LVLM ensemble can be, with ensembles of the same model being much more forgiving if suited to the task at hand.

\paragraph{Our contributions:}
\begin{itemize}
    \item We introduce a unified testbed for training-free LVLM ensembles, which allows fair comparison of approaches along several axes.
    \item We conduct a comprehensive analysis and find that self-assessment of constituent models can be competitive with complex alternatives, even for diverse ensembles.
    \item We find that variables including dataset and ensemble construction play a far more significant role in ensemble performance than previously known, which limits the widespread utility of ensembles.
\end{itemize}

\section{Related Work}
\label{sec:lit_review}

\paragraph{Uncertainty/Confidence Estimation.} This is a very active field, with work spanning fields such as regression \cite{gal_dropout_2016,lakshminarayanan_simple_2017}, classification \cite{jaeger_call_2023}, and segmentation \cite{kendall_what_2017,kahl_values_2024,smith_u-seg_2026}. Confidence measures are typically used for downstream tasks such as failure detection (hallucination detection with LLMs) or calibration \cite{kahl_values_2024,smith_u-seg_2026,agnimo_evaluating_2026,li_evaluating_2023,finlay_caliber_2026} alongside some more specialized cases \eg conformal prediction \cite{urbinati_are_2025}. Our work spans both typical downstream tasks, as the confidence measure is used to detect incorrect predictions but must be calibrated enough for comparison across models. Explicit calibration (\eg temperature scaling \cite{guo_calibration_2017}) is usually assumed to be essential \cite{agrawal_uncertainty-aware_2025}; as we show in \cref{sec:results}, it is possible to do without.

Confidence estimation of a LVLM or LLM response can fall into one of several broad categories \cite{boudiaf_uncertainty-aware_2026}: token-level measures \cite{murray_correcting_2018, fu_deep_2026, kang_scalable_2025,sutskever_sequence_2014,nguyen_probabilities_2026,chen_how_2026,plaut_probabilities_2025,malinin_uncertainty_2021}, verbalized confidence \cite{kadavath_language_2022,wang_calibrating_2026,xi_confidence_2026, li_orce_2026}, response consistency \cite{farquhar_detecting_2024,zhang_consensus_2026,nguyen-mau_enhancing_2025,manakul_selfcheckgpt_2023}, internal state probes \cite{orgad_llms_2025,azaria_internal_2023,radharapu_calibrating_2026}, Bayesian modelling \cite{zhang_fuse_2026}, or a combination \cite{dang_instinct_2026}. In our work, we study token-level or verbalized confidence measures that do not require any training (\cref{sec:app:approaches}). Response consistency and internal state probes either require training, an external embedding model \cite{farquhar_detecting_2024}, or are incompatible with open-ended problems due to simplifying assumptions \cite{cao_clustered_2026}. While supervised uncertainty estimation has proven useful \cite{chauhan_when_2026}, generalization is always a concern. For example, \cite{srey_signals_2026} finds that a lightweight classifier trained on various uncertainty measures, including internal network states, can struggle to generalize to other datasets or domains. Unsupervised uncertainty estimates are also problematic, motivating one of our research questions, as the performance of many measures (token-level, P(True), \etc) is inconsistent in prior work \cite{park_vauq_2026, wang_vlm-uqbench_2026, alazraki_how_2023,yang_calibration_2026, orgad_llms_2025,mi_logical_2026, duan_shifting_2024, tan_basecal_2026, agnimo_evaluating_2026}.

\paragraph{LLM/LVLM Ensembles.}
Aggregating predictions from an ensemble has a long history, including response selection referred to as Dynamic Classifier Selection \cite{zhou_ensemble_2025,ho_decision_1994}. The objective is to maximize performance beyond that of a single model on a wide range of inputs \ie across datasets or distribution shifts. Model routing  \cite{guha_smoothie_2024,lee_confidence-guided_2026} and model selection \cite{zohar_lovm_2023,he_mordal_2026} are sometimes considered a subset of ensembles \cite{chen_harnessing_2026}, where the goal is to be computationally efficient in finding the best model for a task without passing a query to all models. Some work in that domain \cite{guha_smoothie_2024,tan_vision-language_2025,yu_breaking_2024,li_more_2024} could be applied to our testbed, except that they violate at least one of our constraints: they are incompatible with some datasets, require more data than a single model's response, require training or parameter fitting, or would be too computationally expensive to evaluate across all our variables.

While there is some work on open vision language ensembles \cite{tekin_vision_2026,pandey_refine_2025,bugaud_hidden_2026}, most attention in the sequential model domain focuses on (sometimes proprietary) LLMs \cite{chauhan_when_2026,srey_signals_2026, chen_harnessing_2026, jiang_llm-blender_2023}. Ensembles can be used with specialized experts \cite{fan_poly-visual-expert_2024,si_getting_2023}, similar architectures \cite{mai_revisiting_2026,zhang_beyond_2026} or, as in our case, more general models \cite{xia_black-box_2024, yu2026scoop, wang_seeing_2026}, with the expectation that the constituent models are somewhat diverse. Ensembles can also be categorized by the stage at which the samples are merged, with online approaches during inference performing selection \cite{park_ensembling_2025} or aggregation, usually with token-level measures \cite{huang_ensemble_2024,xu_bridging_2024,mavromatis_pack_2024}. Our work falls into the category of offline aggregation after inference, where approaches in line with our requirements (self-assessment, no training or fitted parameters, broadly applicable) are rare; we are aware of two using LLMs \cite{kang_scalable_2025,leang_picsar_2026} and none on LVLMs. A more typical approach makes simplifying assumptions \cite{li_more_2024,yu_cusp_2026,yu2026scoop,tong_diversity_2026,bugaud_hidden_2026,zhao_unleashing_2024,usman_visconf_2026}, is task-specific \cite{wang_seeing_2026,zhang_consensus_2026}, requires training \cite{tekin_vision_2026,pandey_refine_2025,xia_black-box_2024, jiang_llm-blender_2023,bugaud_hidden_2026,lv_urg_2024} or parameter fitting \cite{cho_lookback_2026,zheng_uncertainty-aware_2026,zhang_calibration-aware_2026}.

Ensembles can also be viewed as a form of compute scaling; of note is \cite{chen_scoring_2026}, which creates a peer-review system of models. This is very computationally intensive, making a direct comparison to our testbed impossible to perform. Nonetheless, scaling is known to be effective during training \cite{caballero_broken_2023} and at inference time with \bestofn and related approaches \cite{snell_scaling_2025,huang_cats_2026,rakhsha_majority_2025}, motivating our inclusion of a few \bestofn scenarios.

\section{Preliminaries}

\subsection{Research Questions}
Observation of the literature reveals several open problems around ensembles and LVLMs, which we formulate as research questions:
\begin{enumerate}[label=\textbf{RQ\arabic*}]
    \item Can per-model simple training-free self-assessment approaches be used to effectively ensemble multiple diverse LVLMs? \label{rq:effective-ensemble}
    \item How do such simple approaches compare against existing more complex ones in the literature? \label{rq:compare}
    \item The performance of simple approaches such as mean confidence is inconsistent in prior work. What explains this? \label{rq:variables}
    \item Under what conditions are ensembles of LVLMs useful? Is diversity truly important in practice? \label{rq:useful_diversity}
\end{enumerate}

\subsection{Motivation}
Properly answering these questions motivates the design of our testbed. First, we choose to focus on response \textbf{selection}, as opposed to \textbf{generation} \cite{jiang_llm-blender_2023} where the collected outputs of all constituent models are processed by another model. Our focus on simple approaches and exploring the under-studied variables of dataset choice and ensemble construction means that training a generation model (\eg \cite{tekin_vision_2026,jiang_llm-blender_2023}) is not practical. As well, modern models such as Qwen 3.5 \cite{team_qwen35_2026} are quite capable independently, and thus we expected minimal benefit relative to the complexity of a generation step.

Second, we extend our focus on simplicity to the ranking process by relying on \textbf{model self-assessment}. Existing work often relies on extensive evaluation (\eg cross-response dependencies \cite{chen_scoring_2026,jiang_llm-blender_2023}) where parallelization is difficult. Given the capabilities of modern models and prior work showing that a model's confidence in its answer is useful for failure detection \cite{gerych_who_2024,xu_thinking_2026}, even if miscalibrated \cite{plaut_probabilities_2025}, we hypothesize that we can effectively ensemble models using only each model's confidence estimate as the ranking score for selection. As shown in \cref{fig:meth:testbed}, each response is assessed separately without taking other responses or the question into consideration. This is a natural fit for many confidence estimation approaches, which often assume a single model is being used \cite{orgad_llms_2025, kadavath_language_2022}. Thus, we expect each model to be able to 1) identify its own errors (failure detection) and 2) do so in such a way that its confidence can be compared to that of other models (calibration).

\subsection{Confidence Estimation}
As LVLMs output text, much like LLMs we naturally have access to the probability assigned to each token in the response conditional on all prior tokens. The simplest confidence measure is thus the sequence probability \cite{murray_correcting_2018}:
\begin{equation}
    \label{eq:background:sequence_prob}
    P(t_{1:m}) = \prod_{i=1}^{m} P(t_i \mid t_{1:i-1}),
\end{equation}
where the response is $m$ tokens long and $P(t_i)$ is the probability assigned by the model to the selected (usually the most likely) token at position $i$.

Many other approaches to confidence measurement exist; we explore several dozen ways enumerated in \cref{subsec:variables,sec:app:approaches}. We restrict ourselves to ones that rely on each model's self-assessment, using either token-level confidence or response parsing when the model is asked about its prior answers. This allows parallelization and ensures no training is required, making a more complete exploration of the effect of variables (\eg dataset) possible. This is even more important as many LVLM datasets, including popular ones such as MMMU \cite{yue_mmmu_2024}, do not contain a training set, making it problematic to use any ensemble methods that require training.
\section{Approach}
\label{sec:methodology}

\subsection{Testbed Design}
\label{subsec:testbed_design}
Shown in \cref{fig:meth:testbed}, we introduce a testbed to study ensembles of Large Vision Language Models (LVLMs). We formulate the task as being that of \textbf{candidate response selection}, where all models in a set respond to each question in a dataset and the best response must be chosen as the ensemble's final answer. All models are used as provided by the original authors; no training or fine-tuning is performed.

Formally, we start with a dataset (\labelcref{v:dataset}) that provides a question (text and RGB image) and associated ground truth for evaluation. Questions can be restricted (\eg multiple choice) or open-ended. The text of the question can be amended before being processed by a model depending on the prompt type (\labelcref{v:prompt}). We then have $Q$ LVLMs $\{\mathcal{M}_1, \dots, \mathcal{M}_Q\}$. Each model is generalist and capable of answering all queries independently, but we expect some models to be better than others at some tasks. The set may contain instances of the same model or ones with different architectures (diverse case); the contents of the set are defined by the ensemble configuration (\labelcref{v:ensemble}). Our configurations limit the number of samples: each model is only sampled once in the diverse case, or 3 times in the \bestofn case with a single model. Each model takes as input multimodal data $x$ from the dataset and produces a candidate response $y_i =  \mathcal{M}_i(x)$, where $y_i$ is a text response (tokens and associated probabilities) as expected from a LVLM. This results in a set of candidate responses $\mathbb{Y} = \{y_1, \dots, y_Q\}$.

From this set of candidate responses, we rank them by calculating a confidence score for each response:
\begin{equation}
y_i = \mathcal{M}_i(x), \qquad C_i = f_A(y_i ; \mathcal{M}_i),
\qquad i \in \{1,\dots,Q\},
\end{equation}
where $A$ denotes the approach (\labelcref{v:approach}) chosen, such as P(True) or the sequence probability as described in \cref{sec:app:approaches}. Note that unlike some recent prior work (\eg \cite{jiang_llm-blender_2023}) but in line with some older approaches (\eg \cite{salazar_masked_2020}) there is no direct dependency on the input query $x$. As well, the score assigned to each response is independent of all other responses: each model is expected to evaluate its own response to each question, with no knowledge of the responses of the other models. We thus impose three expectations on the ensembling process: 1) each model is reasonably competent on a wide range of datasets, 2) each model is capable of identifying its own errors more often than not, and 3) each model's self-evaluation is sufficiently well-calibrated (relative to each other) such that we can compare them.

After ranking, we pick the best score (usually the highest, depending on the approach) and evaluate using the typical evaluation process for the selected dataset.

\subsection{Variables}
\label{subsec:variables}
Shown in \cref{fig:meth:testbed}, we examine four different variables with our testbed:
\begin{enumerate}[label=\textbf{V\arabic*}]
    \item \label{v:dataset} \textbf{Dataset}: OCRBench \cite{liu_ocrbench_2024}, MMVet \cite{yu_mm-vet_2024}, MMStar \cite{chen_are_2024}, MMMU \cite{yue_mmmu_2024}, MMBench EN \cite{liu_mmbench_2025}, MathVista \cite{lu_mathvista_2024}, HallusionBench \cite{guan_hallusionbench_2024}, AI2D \cite{kembhavi_diagram_2016}, POPE \cite{li_evaluating_2023}, A-OKVQA \cite{schwenk_-okvqa_2022}, MMMU Pro \cite{yue_mmmu-pro_2025}, OCRVQA \cite{mishra_ocr-vqa_2019}, ScienceQA \cite{lu_learn_2022}, WeMath \cite{qiao_we-math_2025}, LogicVista \cite{xiao_logicvista_2024}, MathVision \cite{wang_measuring_2024}, MMSafetyBench \cite{liu_mm-safetybench_2025}, CMMMU \cite{zhang_cmmmu_2024}, WildVision \cite{lu_wildvision_2024}, MMBench CN \cite{liu_mmbench_2025}. We evaluate on 20 different datasets, covering multiple choice and open-ended questions, OCR, and safety capabilities. Individual LVLMs are expected to be proficient at many tasks, and we want ensembles to not be any more limited than the base models. All datasets are multimodal, containing both images and textual questions and ground truth answers. See \cref{sec:app:dataset_details} for implementation details and a complete list of datasets.
    \item \label{v:prompt} \textbf{Prompt Type}: Normal, Short. Length has long been an issue when calculating confidence in sequential models \cite{murray_correcting_2018}, and modern models remain sensitive to the prompt \cite{liu_understanding_2026}. As such, this variable must be accounted for, and we do so by evaluating both the default prompt for each dataset (\textit{Normal}), and a variant where we request the model to be succinct (\textit{Short}). For the CMMMU and MMSafetyBench datasets, we could not modify the prompt to emphasize succinctness; please see \cref{sec:app:dataset_details} for details and \cref{sec:app:prompt_type} for the prompt wording.
    \item \label{v:ensemble} \textbf{Ensemble Configuration}. \textit{\bestofn}: Qwen 3.5 Instruct \& Thinking, GLM 4.6V Instruct \& Thinking; \textit{Diverse}: Refine and Align, Vision Verification, Modern 1--4, Small Models, Different Size 1--2, Old \& New 1--3, Architecture Variation 1--3, Similar Architecture Variation, Instruct vs. Thinking, Instruct \& Thinking, Top Thinking Performers. We evaluate 24 different ensemble configurations: 4 \bestofn configurations ($N=3$ in all experiments) and 20 more diverse configurations. Diverse predictions are known to be important \cite{dietterich_ensemble_2000}, so it follows that diverse architectures might be useful. The \bestofn scenario uses predictions from multiple runs of the same model. This provides a reference against which any benefits from combining more diverse models can be tested. \bestofn approaches are broadly known to perform well \cite{kang_scalable_2025}, and being the same model are calibrated with respect to each other, bypassing one of the expectations we have on the ensembling process. The more diverse configurations are chosen to ensure sufficient coverage of a wide range of scenarios, including variation in architecture, size, generation mode \etc allowing us to evaluate the utility of different sources of diversity. \cref{tab:app:ensemble_groups} provides a complete list of all configurations and the constituent models used.
    \item \label{v:approach} \textbf{Approach}: Model Score$^\dagger$, Mean Log Confidence, Mean Confidence, Mean Entropy, Sequence Probability$^\dagger$, Maximum Entropy, First \& Last Token Confidence, Minimum Confidence, DC (Deep Confidence)$^\dagger$, P(Confidence), P(True)$^\dagger$, Random. We evaluate 34 different approaches to determining the confidence associated with a model's response, with most plots showing a subset of 24 due to space constraints. Most approaches apply one or more functions to the token-level probabilities assigned by each model, such as the mean, minimum or the confidence in the first token. P(Confidence) prompts each model to identify a key part of its own response, and then uses the token-level confidence of that part. P(True) asks the model if its response was correct. We also evaluate many variants, indicated with $\dagger$; for example, P(True) Search finds a ``True''/``False'' token and uses the assigned token-level confidence rather than relying solely on the text ``True'' or ``False'' in the response. Lastly, the random approach serves as a baseline. Please see \cref{app:tab:approach_list,sec:app:approaches} for the complete list and details on each. In the case of P(True) and P(Confidence), the subsequent queries to the model take advantage of modern caching, ensuring that the initial query is only processed once.
\end{enumerate}

\section{Results}
\label{sec:results}
\begin{figure*}[ht]
    \centering
    \includegraphics[width=0.9\linewidth]{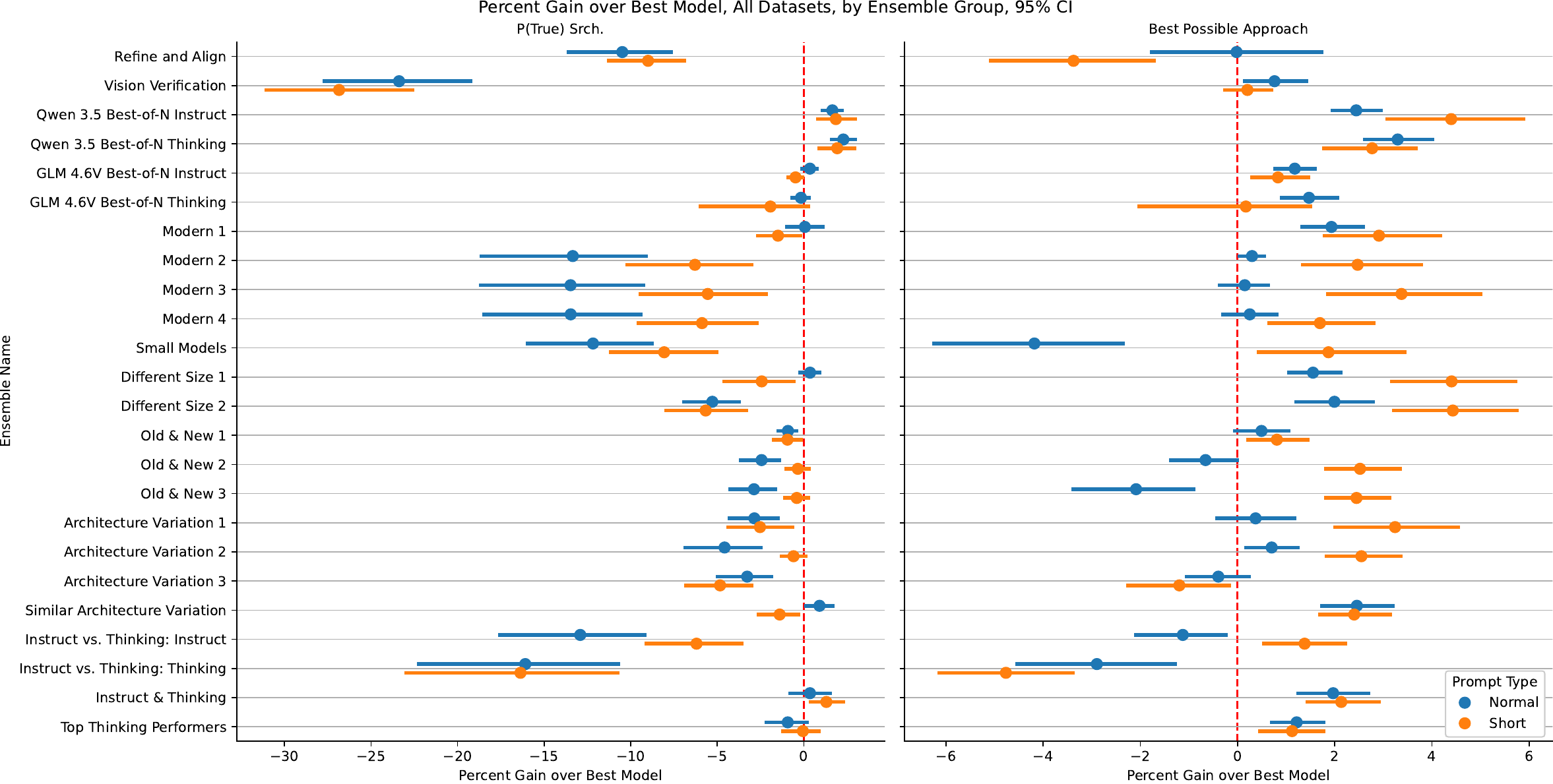}
    \caption{Comparison of different ensembles, presented as the \textbf{Relative Gain (\%)} over the best constituent model, chosen optimally for each dataset, prompt and seed combination. Results show the mean and $95\%$ confidence interval, across all seeds and datasets compatible with the particular ensemble. The confidence interval is calculated using bootstrapping with 2000 iterations. Left: The P(True) Search Approach. Right: Using the optimal approach for each ensemble, dataset and prompt.is calculated using bootstrapping with 2000 iterations.}
    \label{fig:res:percent_gain_best_model}
\end{figure*}
\begin{figure}[ht]
    \centering
    \includegraphics[width=\linewidth]{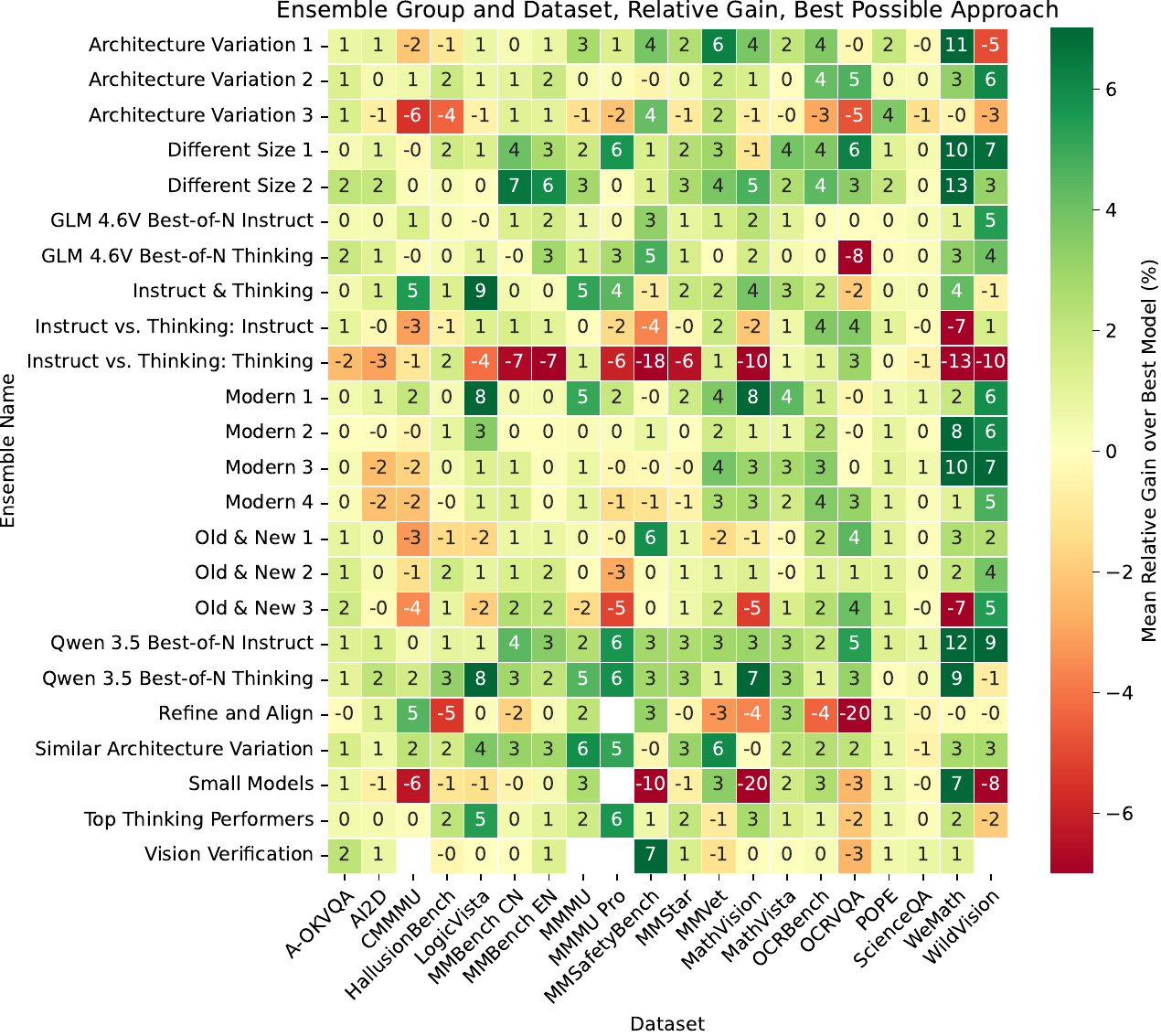}
    \caption{\textbf{Relative Gain (\%)} over the best constituent model, chosen optimally for each dataset, prompt and seed combination. The best possible approach for each ensemble configuration, dataset, and prompt is shown. This shows the best performance that can be obtained compared to an individual model, if we are able to fully characterize the deployment environment. This also serves as a breakdown of the right side of \cref{fig:res:percent_gain_best_model}. Results are averaged across both prompt types (except CMMMU and MMSafetyBench which are only \textit{Normal}) and all seeds.}
    \label{fig:res:heatmap_percent_gain_dataset_best}
\end{figure}
\begin{figure}[ht]
    \centering
    \includegraphics[width=\linewidth]{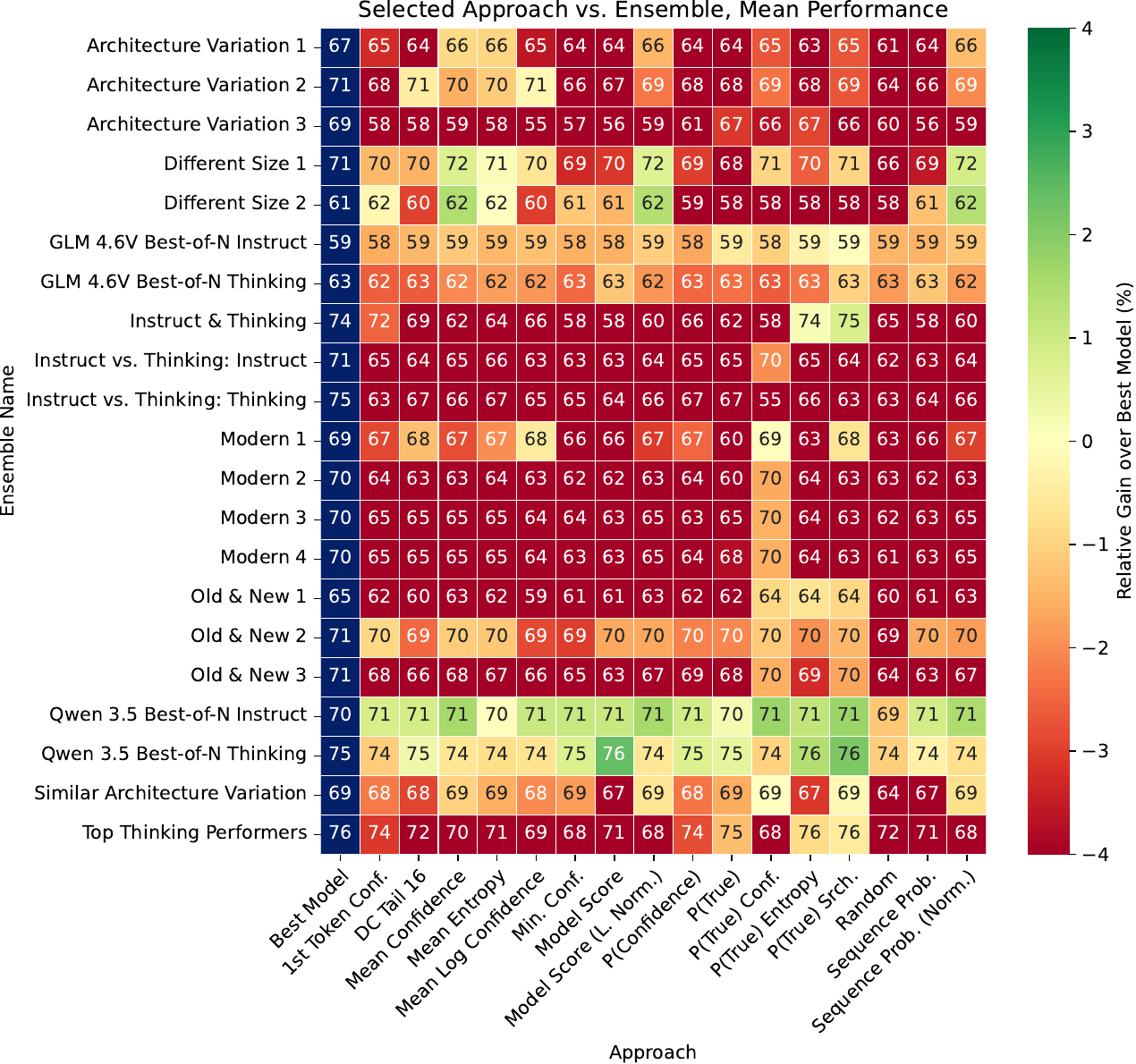}
    \caption{Mean performance of different approaches by ensemble configuration, across all datasets, prompt types and seeds. Green and red shading represents the \textbf{Relative Gain (\%)} over the best constituent model (whose mean performance is shown in blue), chosen optimally for each dataset, prompt and seed combination.}
    \label{fig:res:heatmap_absolute_appr_vs_ensemble}
\end{figure}
\begin{table*}[ht]
\footnotesize
\captionsetup{justification=raggedright,singlelinecheck=false}
\caption{A direct comparison of selected ensembles against their constituent models on several datasets. The table is divided into individual models (top) and ensembles (bottom). ID indicates which individual models make up the ensemble. \textbf{Bold} indicates performance is better than the best possible performance using one of the models that make up the ensemble on that dataset. Values reported as mean ± standard deviation over 3 seeds. Only \textit{Normal} prompts used. Inst. - Instruct, Thnk. - Thinking, BoN - \bestofn, Var. - Variation}
\label{tab:res:direct_ensemble_comparison}
\centering
\begin{adjustbox}{max width=0.9\linewidth}
\begin{tabular}{lllccccc}
\toprule
 &  & Dataset & MMMU & MMSafetyBench & MathVista & OCRBench & WildVision \\
 &  & Metric & Accuracy ↑ & Attack Fail. Rate ↑ & Accuracy ↑ & Score (Norm.) ↑ & Win Rate ↑ \\
ID & Name & Approach &  &  &  &  &  \\
\midrule
0 & Kimi-VL-A3B Inst. & Individual & 52.78 {\scriptsize ±  1.16} & 46.33 {\scriptsize ±  0.81} & 66.60 {\scriptsize ±  0.85} & 87.00 {\scriptsize ±  0.10} & 26.73 {\scriptsize ±  2.50} \\
1 & MiniCPM-V4.5 Inst. & Individual & 61.30 {\scriptsize ±  1.00} & 52.72 {\scriptsize ±  0.50} & 70.20 {\scriptsize ±  1.21} & 86.00 {\scriptsize ±  0.26} & 43.33 {\scriptsize ±  0.99} \\
2 & Ministral-3-8B-2512 Inst. & Individual & 60.07 {\scriptsize ±  0.28} & 58.75 {\scriptsize ±  0.21} & 63.67 {\scriptsize ±  0.32} & 75.93 {\scriptsize ±  0.21} & 44.60 {\scriptsize ±  2.75} \\
3 & GLM-4.6V-Flash Inst. & Individual & 62.52 {\scriptsize ±  1.38} & 37.24 {\scriptsize ±  0.24} & 71.37 {\scriptsize ±  0.15} & 85.67 {\scriptsize ±  0.35} & 42.93 {\scriptsize ±  0.31} \\
4 & InternVL3.5-8B Inst. & Individual & 57.00 {\scriptsize ±  0.67} & 60.22 {\scriptsize ±  1.10} & 66.20 {\scriptsize ±  0.53} & 77.10 {\scriptsize ±  0.96} & 34.00 {\scriptsize ±  0.72} \\
5 & Qwen3.5-9B Inst. & Individual & 72.33 {\scriptsize ±  0.29} & 66.63 {\scriptsize ±  0.41} & 81.80 {\scriptsize ±  0.87} & 84.03 {\scriptsize ±  0.21} & 57.47 {\scriptsize ±  2.14} \\
6 & Qwen3.5-9B Thnk. & Individual & 67.70 {\scriptsize ±  0.45} & 67.68 {\scriptsize ±  0.47} & 81.70 {\scriptsize ±  0.66} & 87.93 {\scriptsize ±  0.68} & 55.47 {\scriptsize ±  1.90} \\
7 & Qwen3.5-9B Thnk.-BoN & Individual & 68.37 {\scriptsize ±  0.53} & 67.46 {\scriptsize ±  0.64} & 81.70 {\scriptsize ±  0.56} & 88.07 {\scriptsize ±  0.65} & 54.00 {\scriptsize ±  1.06} \\
8 & Qwen3.5-9B Thnk.-BoN & Individual & 67.56 {\scriptsize ±  0.48} & 67.84 {\scriptsize ±  0.51} & 80.93 {\scriptsize ±  0.60} & 87.97 {\scriptsize ±  0.81} & 54.93 {\scriptsize ±  1.17} \\
9 & Qwen3.5-9B Thnk.-BoN & Individual & 67.56 {\scriptsize ±  0.69} & 67.32 {\scriptsize ±  0.53} & 81.60 {\scriptsize ±  0.53} & 87.30 {\scriptsize ±  0.36} & 54.93 {\scriptsize ±  2.16} \\
\cline{1-8} \cline{2-8}
\multirow[c]{3}{*}{0,1,2,4} & \multirow[c]{3}{*}{Architecture Var. 1} & 1st Token Conf. & 58.52 {\scriptsize ±  0.67} & 57.14 {\scriptsize ±  0.49} & 69.90 {\scriptsize ±  0.26} & \bfseries 87.23 {\scriptsize ±  0.55} & 41.07 {\scriptsize ±  1.62} \\
 &  & P(True) Conf. & \bfseries 61.89 {\scriptsize ±  0.97} & 56.55 {\scriptsize ±  0.45} & \bfseries 70.93 {\scriptsize ±  0.40} & \bfseries 88.03 {\scriptsize ±  0.51} & 41.80 {\scriptsize ±  0.53} \\
 &  & P(True) Srch. & \bfseries 61.93 {\scriptsize ±  1.10} & 56.25 {\scriptsize ±  0.00} & \bfseries 70.87 {\scriptsize ±  0.47} & \bfseries 88.03 {\scriptsize ±  0.51} & 41.07 {\scriptsize ±  2.23} \\
\cline{1-8} \cline{2-8}
\multirow[c]{3}{*}{3,5} & \multirow[c]{3}{*}{Modern 2} & 1st Token Conf. & 65.04 {\scriptsize ±  1.52} & 60.18 {\scriptsize ±  0.53} & 77.03 {\scriptsize ±  0.65} & \bfseries 87.70 {\scriptsize ±  0.17} & 51.73 {\scriptsize ±  0.50} \\
 &  & P(True) Conf. & \bfseries 72.52 {\scriptsize ±  0.23} & \bfseries 67.16 {\scriptsize ±  0.41} & \bfseries 81.83 {\scriptsize ±  0.85} & 84.03 {\scriptsize ±  0.21} & \bfseries 57.73 {\scriptsize ±  3.10} \\
 &  & P(True) Srch. & 66.74 {\scriptsize ±  0.86} & 56.35 {\scriptsize ±  0.19} & 76.07 {\scriptsize ±  0.78} & \bfseries 86.67 {\scriptsize ±  0.25} & 46.27 {\scriptsize ±  0.76} \\
\cline{1-8} \cline{2-8}
\multirow[c]{3}{*}{3,6} & \multirow[c]{3}{*}{Inst. \& Thnk.} & 1st Token Conf. & 67.59 {\scriptsize ±  0.34} & 67.12 {\scriptsize ±  0.33} & \bfseries 81.70 {\scriptsize ±  0.70} & 87.90 {\scriptsize ±  0.66} & 54.53 {\scriptsize ±  0.61} \\
 &  & P(True) Conf. & 62.41 {\scriptsize ±  1.39} & 37.56 {\scriptsize ±  0.60} & 71.37 {\scriptsize ±  0.15} & 85.67 {\scriptsize ±  0.35} & 44.67 {\scriptsize ±  1.17} \\
 &  & P(True) Srch. & \bfseries 70.67 {\scriptsize ±  0.44} & 63.97 {\scriptsize ±  0.03} & \bfseries 83.83 {\scriptsize ±  0.59} & \bfseries 88.87 {\scriptsize ±  0.61} & 53.07 {\scriptsize ±  2.00} \\
\cline{1-8} \cline{2-8}
\multirow[c]{3}{*}{7,8,9} & \multirow[c]{3}{*}{Qwen 3.5 BoN Thnk.} & 1st Token Conf. & 67.89 {\scriptsize ±  1.44} & \bfseries 68.00 {\scriptsize ±  0.27} & 81.37 {\scriptsize ±  0.78} & 87.67 {\scriptsize ±  0.71} & 54.53 {\scriptsize ±  0.46} \\
 &  & P(True) Conf. & 68.07 {\scriptsize ±  1.13} & 67.82 {\scriptsize ±  0.12} & 81.50 {\scriptsize ±  1.39} & 87.80 {\scriptsize ±  0.40} & 53.93 {\scriptsize ±  1.22} \\
 &  & P(True) Srch. & \bfseries 70.00 {\scriptsize ±  0.51} & \bfseries 68.53 {\scriptsize ±  1.16} & \bfseries 84.13 {\scriptsize ±  0.46} & \bfseries 89.37 {\scriptsize ±  0.76} & \bfseries 57.67 {\scriptsize ±  2.32} \\
\bottomrule
\end{tabular}

\end{adjustbox}
\end{table*}
\begin{table*}[ht]
\footnotesize
\captionsetup{justification=raggedright,singlelinecheck=false}
\caption{Comparison against \cite{pandey_refine_2025, tekin_vision_2026}. Results represent the \textbf{Relative Gain (\%)} over the best constituent model, chosen optimally for each dataset, with mean ± standard deviation over 3 seeds. Only \textit{Normal} prompts used. \textbf{Bold} highlights positive values. Missing values are due to incompatibility or different choices in dataset selection and evaluation compared to prior work. ``Approx.'' indicates the closest ensemble we could build to match; see ``Refine and Align'' and ``Vision Verification'' in \cref{tab:app:ensemble_groups}. For OCRVQA, \cite{tekin_vision_2026} uses F1 while we use accuracy.}
\label{tab:res:refine_verfication_compare}
\centering
\begin{adjustbox}{max width=0.9\linewidth}
\begin{tabular}{llcccccc}
\toprule
 & Dataset & VQARad & ScienceQA & A-OKVQA & MMMU & MMMU Pro & OCRVQA \\
 & Metric & Acc. ↑ & Acc. ↑ & Acc. ↑ & Acc. ↑ & Acc. ↑ & F1/Acc. ↑ \\
Ensemble & Approach &  &  &  &  &  &  \\
\midrule
\multirow[c]{3}{*}{\cite{pandey_refine_2025}} & Agentic+AlignCal+FL & -1.16 & \bfseries 0.13 & - & - & - & - \\
 & Agentic+FL & -0.72 & -2.11 & - & - & - & - \\
 & Agentic+LS & -1.88 & -1.05 & - & - & - & - \\
\cline{1-8}
\multirow[c]{2}{*}{\cite{tekin_vision_2026}} & V3Fusion-LED & - & - & \bfseries 0.71 & -0.48 & -1.53 & \bfseries 2.38 \\
 & V3Fusion-Rectify & - & - & \bfseries 2.21 & \bfseries 8.81 & \bfseries 4.87 & \bfseries 2.09 \\
\cline{1-8}
\multirow[c]{5}{*}{Approx. \cite{pandey_refine_2025}} & DC Bott. 10\%, W. 8 & - & -1.11 {\scriptsize ±  0.09} & \bfseries 0.35 {\scriptsize ±  0.05} & \bfseries 2.58 {\scriptsize ±  1.22} & - & -50.82 {\scriptsize ±  0.13} \\
 & DC Tail 10\% & - & -0.27 {\scriptsize ±  0.21} & \bfseries 0.00 {\scriptsize ±  0.00} & \bfseries 1.25 {\scriptsize ±  1.26} & - & -47.65 {\scriptsize ±  0.16} \\
 & Max. Entropy & - & \bfseries 0.00 {\scriptsize ±  0.05} & -0.35 {\scriptsize ±  0.05} & \bfseries 4.64 {\scriptsize ±  0.39} & - & -24.98 {\scriptsize ±  0.01} \\
 & Model Score (L. Norm.) & - & -1.76 {\scriptsize ±  0.23} & -0.57 {\scriptsize ±  0.10} & \bfseries 3.02 {\scriptsize ±  2.01} & - & -21.43 {\scriptsize ±  0.02} \\
 & P(True) Srch. & - & -8.57 {\scriptsize ±  0.28} & -4.06 {\scriptsize ±  0.24} & -2.43 {\scriptsize ±  1.22} & - & -32.50 {\scriptsize ±  0.11} \\
\cline{1-8}
\multirow[c]{5}{*}{Approx. \cite{tekin_vision_2026}} & DC Bott. 10\%, W. 8 & - & -0.69 {\scriptsize ±  0.26} & -0.81 {\scriptsize ±  0.30} & - & - & -8.66 {\scriptsize ±  0.09} \\
 & DC Tail 10\% & - & -0.36 {\scriptsize ±  0.20} & -4.97 {\scriptsize ±  0.47} & - & - & -9.52 {\scriptsize ±  0.09} \\
 & Max. Entropy & - & \bfseries 0.35 {\scriptsize ±  0.05} & \bfseries 1.65 {\scriptsize ±  0.41} & - & - & -5.40 {\scriptsize ±  0.03} \\
 & Model Score (L. Norm.) & - & \bfseries 0.93 {\scriptsize ±  0.30} & \bfseries 2.55 {\scriptsize ±  0.42} & - & - & -3.01 {\scriptsize ±  0.04} \\
 & P(True) Srch. & - & -16.35 {\scriptsize ±  0.12} & -2.08 {\scriptsize ±  1.01} & - & - & -31.29 {\scriptsize ±  0.08} \\
\bottomrule
\end{tabular}

\end{adjustbox}
\end{table*}

In \cref{fig:res:percent_gain_best_model,fig:res:heatmap_percent_gain_dataset_best,tab:res:refine_verfication_compare} we use \textit{Relative Gain}, where a given ensemble's performance is measured relative to using the best constituent ensemble model for a given dataset, prompt and seed combination. A value of 0\% is equivalent to being able to dynamically select the best single model (of those used in the ensemble) for each case. This metric has been used in prior work \cite{tekin_vision_2026} and answers the question: can the ensemble effectively select from the available predictions? It is not a theoretical maximum, where the approach would always be able to identify a correct prediction from a larger set, but is a more reasonable baseline that we expect an ensemble to match. Finally, \cref{tab:res:direct_ensemble_comparison,fig:res:heatmap_absolute_appr_vs_ensemble} forgo this relative measure and instead use each dataset's metric directly or averaged respectively. This gives a more intuitive view as to how each dataset, approach and ensemble can perform.

\subsection{Analysis}
\label{subsec:res:analysis}
\paragraph{Ensemble composition, dataset and prompt all have an impact; joint consideration is required:} In \cref{fig:res:heatmap_absolute_appr_vs_ensemble}, an approach that results in a positive relative gain may result in a loss with a different ensemble. Meanwhile, \cref{fig:res:percent_gain_best_model} shows that if we use a ``good'' (see \cref{fig:app:relax_win_rate_plot}) overall approach (on the left) a different one (shown on the right) can often surpass it. Adding an instruction to each model to be succinct also has an impact; shorter responses allow higher relative gains for token-level aggregation approaches, matching some recently observed trends \cite{seifi_dissecting_2026}. If we use the optimal approach, shown in  \cref{fig:res:heatmap_percent_gain_dataset_best}, good performance is broadly achievable albeit fragile.

Putting it all together, the picture that emerges is that the ensemble configuration, dataset choice and prompt specification \textbf{must be jointly optimized}. Optimizing one variable at a time makes it easy to fall into a local minima.
 
\paragraph{Model characteristics are a driving factor:} Model-specific details have a clear effect on ensemble performance. As shown in \cref{fig:res:percent_gain_best_model,fig:res:heatmap_percent_gain_dataset_best}, ensembles composed of older (\eg \textit{Refine and Align}) or smaller models perform relatively worse than more modern combinations. Configurations that use thinking model also pose challenges, particularly for approaches like P(True) Confidence which are not designed to handle them. Overall, however, many configuration strategies, ranging from \bestofn to much more diverse combinations are able to achieve a positive gain and there do not appear to be any configuration strategies (\eg prioritizing architecture diversity or different sizes of the same model) that are complete failures. What we do see in \cref{fig:res:heatmap_absolute_appr_vs_ensemble}, which looks at absolute instead of relative performance, is that some models are much more difficult to ensemble than others \eg GLM 4.6V, used in ``Modern 1-4'' and in a \bestofn configuration. In those cases, where P(True) inadvertently avoids selecting the GLM model (see \cref{sec:app:p_true_spurious}) performance becomes acceptable.

\paragraph{Many ensemble configurations are useful:} In \cref{tab:res:direct_ensemble_comparison}, we directly compare various ensembles on selected datasets against their constituent models. We see that it is usually possible to find one approach that is able to effectively combine the predictions of the constituent models for most of the shown datasets, performing better than the best individual model. This includes ensembles of the same architecture (\textit{\bestofn}) as well as more diverse ones (\textit{Architecture Variation}). However, in line with our other results there are many cases where an approach that works well with one ensemble or dataset does not work well with another.

\paragraph{Best-of-$\boldsymbol{N}$ is a good all-rounder:} One final trend is that \bestofn ensembles are one of the easiest to work with. This is especially apparent in \cref{fig:res:heatmap_absolute_appr_vs_ensemble,fig:res:heatmap_percent_gain_dataset_best} where across datasets and approaches many configurations work reasonably well when compared to more diverse cases. This is not surprising, as using only a single model avoids calibration issues when comparing scores from different models. The Qwen 3.5 model also helps by providing good absolute performance (\cref{tab:app:individual_mod_results}), minimizing worst-case results in the \bestofn scenario. This trend is in contrast to \cite{tong_diversity_2026,bugaud_hidden_2026} who find model diversity valuable, but assume answers can be grouped \eg via multiple choice questions.

\subsection{Comparison against prior work}
\label{subsec:res:prior_work_compare}
\cref{tab:res:refine_verfication_compare} compares selected approaches against prior work \cite{tekin_vision_2026, pandey_refine_2025} on a set of common datasets. Unfortunately, we are unable to provide a one-to-one comparison against the original work; see \cref{sec:app:comparison_issues} for details.
We use the relative gain metric as with \cref{fig:res:percent_gain_best_model,fig:res:heatmap_percent_gain_dataset_best}.

\paragraph{Token-level confidence can be competitive with training:} The main finding from \cref{tab:res:refine_verfication_compare} is that simple approaches that use token-level information generated by the model can be competitive with more complex approaches that require training or setting parameters. This applies when we have similar ensembles and is especially true against \textit{V3Fusion-LED} or \textit{Agentic+FL}, which omit some parts of the complete approaches in \cite{pandey_refine_2025,tekin_vision_2026} but still require training data. However, the high performance variation observed elsewhere continues to apply, with all shown approaches struggling on the OCRVQA dataset and often generalizing poorly.

\paragraph{Secondary aspects are critical:} All of the approaches we evaluate are based on sequential model fundamentals or observations about their behaviour that have been proven in some related context, such as the ability to correctly infer the most important word in a response \cite{duan_shifting_2024,orgad_llms_2025}. However, we find that many approaches struggle when used outside of the original framework.

For example, \textit{Deep Confidence} was shown to be better than \textit{Mean Log Confidence} across lengthy reasoning traces of a single model \cite{fu_deep_2026}. However, in \cref{fig:res:heatmap_absolute_appr_vs_ensemble} we see this claim holds only in contexts similar to those of \cite{fu_deep_2026} \ie Thinking models. As another example, \cite{orgad_llms_2025} show that selected LLMs are able to extract key tokens from their own responses, and that ``truthfulness information is concentrated in the exact answer tokens''. Our results (\cref{fig:res:heatmap_absolute_appr_vs_ensemble}) show that across a large set of models and datasets, the key tokens alone are not a leading approach. If we compare our results with Tab. 1 in \cite{orgad_llms_2025}, it becomes clear that the key tokens are most useful when used in conjunction with a trained probe tailored to a specific model. As we do not implement this step, the location of key tokens is only of some limited utility.

For additional results, including the performance of the models used in the ensembles by themselves and a more detailed breakdown of the results shown here, please see \cref{sec:app:approach_compare,sec:app:additional_findings,sec:app:more_ensemble_compare,sec:app:individual_results}.

\subsection{Discussion}
\label{subsec:discussion}
From our results, we can answer our Research Questions:

\noindent \ref{rq:effective-ensemble}. \textbf{It is possible to effectively ensemble even diverse models using each model's own confidence.} \cref{fig:res:percent_gain_best_model} shows that it is usually possible to perform better than the best individual model, with \cref{tab:res:direct_ensemble_comparison} reinforcing this point with specific examples. Prior work concluded that ``LVLM sequence probabilities are not reliable enough to be used for ensembling'' and that P(True) ``may not be a reliable method'' \cite{alazraki_how_2023}, but our work shows this to not be true overall. In fact, P(True) variants are some of the best approaches we tested.

\noindent \ref{rq:compare}. \textbf{Training-free approaches compatible with open-ended tasks can be competitive with more complex alternatives.} Following \cite{alazraki_how_2023}, several ensembling methods for vision language models have been proposed \cite{xia_black-box_2024,pandey_refine_2025,tekin_vision_2026,zhang_why_2026}, all of which require training/parameter fitting, are limited by a choice of embedding \cite{farquhar_detecting_2024}, \etc. \cite{agrawal_uncertainty-aware_2025} proposes a training-free open-ended compatible approach using token-level metrics, but requires each response to be processed by every model to approximate relative calibration. As well, modern LVLMs are highly capable across a wide range of tasks, yet several ensemble approaches (\eg \cite{yu2026scoop,li_more_2024}) remain restricted to multiple choice questions, with one \cite{tong_diversity_2026} showing that CoT (\textit{chain-of-thought}) prompts are an additional requirement. We show in \cref{subsec:res:prior_work_compare} that simpler token-level or self-reflection methods can be competitive without requirements such as CoT prompts, training or calibration.

\noindent \ref{rq:variables}. \textbf{Ensemble composition, dataset, and other variables must be accounted for.} In \cref{subsec:res:analysis}, we note that the ensemble composition, prompt type and dataset choice all significantly impact the potential performance benefits of an ensemble. We also find that secondary aspects (\cref{subsec:res:analysis}) and small details (\cref{sec:app:additional_findings}) are critical. However, prior work (\eg \cite{pandey_refine_2025, tekin_vision_2026,park_vauq_2026, wang_vlm-uqbench_2026}) perform little to no exploration of these variables, explaining the inconsistent baseline results and making it unclear how well any given approach may generalize (particularly important for LVLMs). Our suggestion is that any proposed ensemble approach involving large vision language models, and intended to be general purpose, must go through a thorough evaluation encompassing the ensemble composition, prompt specification and dataset. Baseline methods should also receive at least some attention so as to not disadvantage them.

\noindent \ref{rq:useful_diversity}. \textbf{Diversity is difficult to exploit; a requirement to jointly optimize many variables undermines the utility of LVLM ensembles.} Ensembles are used when better performance is needed than possible from individual models \cite{zhou_ensemble_2025}, and LVLMs are capable of handling a wide variety of open-ended tasks. As we show in \cref{tab:res:direct_ensemble_comparison}, with ensembles we want to --- and can --- effectively leverage diversity to do better across such datasets. However, as we discuss in \cref{subsec:res:analysis} there is a need to jointly optimize for several variables such as ensemble composition and dataset. In other words, to get the most out of an ensemble on several datasets we need to optimize based on the constituent models and the datasets, and outside of those conditions we may perform worse than a good single model. This is especially true for configurations of diverse models (different architectures, sizes, \etc) where only a few approaches may result in a positive relative gain. High relative gains are seen only when making strong assumptions or with complex specially optimized approaches \cite{tong_diversity_2026, wang_seeing_2026}. This makes diverse ensembles difficult to use. As such, our recommendation to practitioners is to first explore a \bestofn approach with a strong model such as the Qwen series, as its worst-case performance is still that of a good single model. A diverse ensemble should only be used if deployment conditions are constrained and well understood. If training a model is an option, with known deployment conditions it may also be worth considering fine-tuning a single model and avoiding the inference time cost of ensembles.

\section{Conclusion}
In this work, we introduce a testbed allowing for a systematic study of ensemble methods for Large Vision Language Models. We use this testbed to answer several under-explored research questions, identifying several novel findings. First, simple confidence measures that rely on the self-assessment of constituent models can be enough to be competitive with more complex alternatives when it comes to ensembles, even with diverse constituent models. Second, dataset choice, ensemble composition and other variables play a significant role in ensemble performance, which can overshadow that of the approach used. This leads us to the conclusion that while diverse LVLM ensembles can serve a purpose, their generalizability is limited, which undermines their general utility. For practitioners, diverse training-free ensembles can work if deployment conditions are well-defined, but otherwise a \bestofn approach with a strong model is simple and often effective. Researchers intending to use ensembles (diverse or \bestofn) should first consider simpler methods, such as those that use token-level confidence, before investigating more complex alternatives \eg those that require training. 

\section{Acknowledgements}
NVIDIA provided one of the GPUs used during development. Most experiments were made possible by support provided by Calcul Québec, Compute Ontario, the BC DRI Group, Prairies DRI, and the Digital Research Alliance of Canada.

\section{LLM Usage Statement}
LLMs were used for some brainstorming, the generation of some code for experiments, as a supplement to a manual literature search, to assist with the generation of some tables and equations including \cref{app:tab:dataset_map,tab:app:ensemble_groups,app:tab:approach_list,eq:app:conf_p_true,tab:app:sampling_params}, and to help with listing all datasets in \cref{sec:app:dataset_details}. LLMs were also used to assist with some \LaTeX\xspace formatting issues, identify typos, and to act as a preliminary reviewer to identify potential deficiencies before submission. All text, figures and tables not explicitly mentioned were created by hand. All content was carefully reviewed and manually edited for correctness or clarity, as the agents were found to sometimes introduce errors or irrelevant information. The authors assume full responsibility for the content of this paper.

Available models and tools changed frequently during the creation of this paper, with new releases within months of each other, and as such the latest model available at any given time was used. Proprietary models belonging to the GPT series (via Microsoft Copilot) were used in a chat configuration, with the most recent model being GPT 5.6 Sol. Open-weight models (GLM 4.7 Flash, Qwen 3.5-3.8 27B/35B-A3B, DeepSeek V4 Flash 0731) were used in a chat configuration via Open WebUI or in an agentic configuration via the Pi harness for general work. Roo Code and subsequently Kilo Code were used as the agentic harness for specific coding tasks within an IDE.
{
    \small
    \bibliographystyle{ieeenat_fullname}
    \bibliography{main}
}

\clearpage
\appendix
\setcounter{page}{1}
\renewcommand\thesection{A\arabic{section}}
\renewcommand\thefigure{A\arabic{figure}}
\renewcommand\thetable{A\arabic{table}}
\renewcommand\thepage{A-\arabic{page}}
\setcounter{figure}{0}
\setcounter{table}{0}
\maketitlesupplementary

\section{Datasets and metrics}
\label{sec:app:dataset_details}
This work uses the VLMEvalKit \cite{duan_vlmevalkit_2024} framework, which provides a unified implementation of all 20 datasets used. In \cref{app:tab:dataset_map}, we provide a mapping from the dataset names we refer to and the corresponding VLMEvalKit identifiers that uniquely identify them, as there are multiple dataset revisions and splits in several cases. We also list the type of questions and the number of questions. For datasets where multiple splits exist (\eg \texttt{dev} and \texttt{val}), the \texttt{val} (validation) split is used. Only splits that contain ground truth labels for evaluation are used; any splits where evaluation can only be performed using an external server are discarded. 

For each dataset, we select one key evaluation metric proposed by the dataset authors related to overall performance. This is usually but not always accuracy; please see the end of this section for a review and the original work for further details. Criteria and protocols around evaluation follow the dataset procedure as implemented by VLMEvalKit, which in some cases includes a revised protocol from the original publication. We break from VLMEvalKit standard procedure however with respect to how each model is prompted to answer a question from a dataset: all models are provided the same prompt for a given dataset. In other words, dataset prompts are not optimized for each model. Evaluation of a model's prediction is done either deterministically via a heuristic (\eg string matching) or stochastically via LLM-as-a-judge following the default VLMEvalKit implementation. We harmonize all judge evaluation to use OpenAI's GPT-OSS 120B, instead of the variety of out-of-date proprietary models originally called for. Some judge prompts are modified from the dataset default so as to improve clarity, robustness and compatibility with recent models. We observed that using the default prompts for some datasets with the GPT-OSS series as a judge would result in responses such as ``You didn't ask a question''. This ``error'' is correct, as the prompt did not in fact ask a question, being formatted in the style of a completion compatible with older models but confusing even to a human reader. Please see \cref{sec:app:judge_prompts} for the exact judge prompts used.

Due to one or more of several factors, some figures and tables have missing data, including \cref{fig:res:heatmap_percent_gain_dataset_best,tab:app:individual_mod_results,tab:app:abs_ensemble_vs_dataset_best_approach}. In some cases, incompatibilities existed between the dataset and another parameter, such as the prompt type or the model used. In the case of CMMMU \cite{zhang_cmmmu_2024}, we could not modify the prompt to instruct the model to be succinct as we lacked the requisite language proficiency. In the case of MMSafetyBench, instructing the model to be succinct would interfere with the evaluation structure, where multiple variants of a given prompt exist with some explicitly calling for detailed steps. Other cases of missing results are due to a model being unable to process one or more questions in the dataset due to errors beyond our control such as maximum context length, the maximum number of images, \etc. This is more common on older models, with newer ones not suffering from the same limitations. We do not report results for any datasets where a model was only capable of processing a subset of questions. A model responding but hitting the maximum output length \eg due to entering an infinite loop is not considered an error, with evaluation proceeding as normal. We note that in some cases, we encountered issues with some model and dataset combinations that others did not (\eg the LLaVa series and the MMMU-Pro dataset used by \cite{tekin_vision_2026}). This is likely implementation related, as we use VLLM \cite{kwon_efficient_2023} and \cite{tekin_vision_2026} uses the Transformers library \cite{wolf_transformers_2020} with custom settings for each model. We consider debugging complex software packages or introducing custom implementations for a minority of models to be beyond our scope.

We are aware of three limitations around our dataset choices. First, we do not correct any errors or otherwise improve the ground truth. During development, we noted several cases where the model was correct and the ground truth was incorrect. We also found several instances of ambiguous questions or poorly formatted ground truth answers where the correctness of an answer could be subjective. Consider the single example in \cref{fig:meth:testbed}, the selected responses from our experiments and the provided ground truth. Is the answer ``above'' correct by itself? What about ``upper left of man''? Such subjectivity is usually decided by a LLM-as-a-judge.

Second, we ignore the source(s) used for each dataset. Thus, some questions may exist in more than one dataset. One example is MMStar, which contains curated questions from MMBench, AI2D, SEEDBench, MathVista, and ScienceQA. Another is MMBench, which includes questions sourced from the general internet, the COCO training set \cite{lin_microsoft_2014}, online courses, the ScienceQA dataset, \etc. De-duplicating datasets is beyond our scope; overlap between datasets is not ideal, but all are standard benchmarks and exhibit sufficiently different performance from each other.

Third, it is possible that one or more evaluated models may have seen some of the questions during training. As we are not in control of and have limited visibility into the training of many models, we cannot address this issue beyond assuming that the original authors took appropriate precautions to avoid data contamination on known standard datasets, which includes all datasets used in this work.

\begin{table*}
\centering
\caption{Datasets used in this work, their identifiers in VLMEvalKit~\cite{duan_vlmevalkit_2024}, their question types, and the number of questions evaluated. The number of questions may be different from total published dataset size as we use only splits with public ground truth data. \textit{Mixed VQA} indicates a combination of multiple choice and open-ended VQA. \textit{Safety} is unique; see \cite{liu_mm-safetybench_2025}. Metrics are discussed in \cref{sec:app:dataset_details}.}
\label{app:tab:dataset_map}
    \begin{tabular}{lllc}
    \toprule
    Dataset Name & VLMEvalKit Identifier & Type & Questions \\
    \midrule
    OCRBench \cite{liu_ocrbench_2024} & OCRBench & OCR & 1000 \\
    MMVet \cite{yu_mm-vet_2024} & MMVet & Open-ended VQA & 218 \\
    MMStar \cite{chen_are_2024} & MMStar & Multiple choice VQA & 1500 \\
    MMMU \cite{yue_mmmu_2024} & MMMU\_DEV\_VAL & Mixed VQA & 900 \\
    MMBench EN \cite{liu_mmbench_2025} & MMBench\_DEV\_EN\_V11 & Multiple choice VQA & 1164 \\
    MathVista \cite{lu_mathvista_2024} & MathVista\_MINI & Mixed Math VQA & 1000 \\
    HallusionBench \cite{guan_hallusionbench_2024} & HallusionBench & Multiple choice (yes/no) & 951 \\
    AI2D \cite{kembhavi_diagram_2016} & AI2D\_TEST & Multiple choice VQA & 3088 \\
    POPE \cite{li_evaluating_2023} & POPE & Multiple choice (yes/no) & 5127 \\
    A-OKVQA \cite{schwenk_-okvqa_2022} & A-OKVQA & Multiple choice VQA & 1145 \\
    MMMU Pro \cite{yue_mmmu-pro_2025} & MMMU\_Pro\_10c & Multiple choice VQA & 1730 \\
    OCRVQA \cite{mishra_ocr-vqa_2019} & OCRVQA\_TEST & OCR & 99929 \\
    ScienceQA  \cite{lu_learn_2022} & ScienceQA\_VAL & Multiple choice VQA & 2097 \\
    WeMath  \cite{qiao_we-math_2025} & WeMath & Multiple choice Math VQA & 1740 \\
    LogicVista  \cite{xiao_logicvista_2024} & LogicVista & Multiple choice VQA & 447 \\
    MathVision \cite{wang_measuring_2024} & MathVision & Mixed Math VQA & 3040 \\
    MMSafetyBench \cite{liu_mm-safetybench_2025} & MMSafetyBench & Safety & 1680 \\
    CMMMU \cite{zhang_cmmmu_2024} & CMMMU\_VAL & Mixed VQA & 900 \\
    WildVision \cite{lu_wildvision_2024} & WildVision & Open-ended VQA & 500 \\
    MMBench CN \cite{liu_mmbench_2025} & MMBench\_DEV\_CN\_V11 & Multiple choice VQA & 1164 \\
    \bottomrule
    \end{tabular}
\end{table*}

We finish with a list of all datasets used, along with the corresponding metric and a brief summary of the task and evaluation process if unique:
\begin{itemize}
    \item \textbf{OCRBench.} \cite{liu_ocrbench_2024} \textit{Score (Normalized).}
    Each question asks for the text in the entire image or a portion of it in the case of more complex images such as those with tables. The score includes the model's performance on several sub tasks, such as text recognition and knowledge reasoning. 

    \item \textbf{MMVet.} \cite{yu_mm-vet_2024} \textit{Accuracy.}
    Open-ended questions across several categories including math, OCR, and recognition.

    \item \textbf{MMStar.} \cite{chen_are_2024} \textit{Accuracy.}
    Multiple choice questions, curated from several other datasets.

    \item \textbf{MMMU.} \cite{yue_mmmu_2024} \textit{Accuracy.}
    Mostly multiple choice and open-ended questions across various fields.

    \item \textbf{MMBench EN.} \cite{liu_mmbench_2025} \textit{Accuracy (circular).}
    Multiple choice questions spanning multiple categories, with circular evaluation which creates permutations of the 1164 unique questions with the ordering of the options randomized. A question is correct only if all (usually 4) permutations are correct.

    \item \textbf{MathVista.} \cite{lu_mathvista_2024} \textit{Accuracy.}
    Multiple choice and open-ended questions focused on math.

    \item \textbf{HallusionBench.} \cite{guan_hallusionbench_2024} \textit{Average accuracy (aAcc).}
    Binary questions about image contents. Note that the dataset has 1129 questions but some are text-only and are skipped.

    \item \textbf{AI2D.} \cite{kembhavi_diagram_2016} \textit{Accuracy.}
    Multiple choice questions about elementary-level diagrams.

    \item \textbf{POPE.} \cite{li_evaluating_2023} \textit{F1 Score.}
    Binary questions about the presence of objects in images.

    \item \textbf{A-OKVQA.} \cite{schwenk_-okvqa_2022} \textit{Accuracy.}
    Multiple choice questions where common sense reasoning is required.

    \item \textbf{MMMU Pro.} \cite{yue_mmmu-pro_2025} \textit{Accuracy.}
    A filtered and augmented version of MMMU which is much more challenging yet retains the multiple choice format.

    \item \textbf{OCRVQA.} \cite{mishra_ocr-vqa_2019} \textit{Accuracy.}
    Questions that require reading text in images and some reasoning \eg identifying authors or the category of an item.

    \item \textbf{ScienceQA.} \cite{lu_learn_2022} \textit{Accuracy.}
    Multiple-choice science questions.

    \item \textbf{WeMath.} \cite{qiao_we-math_2025} \textit{Score (Strict).}
    1740 math problems are grouped into 525 composite groups. The Strict Score metric evaluates the composite groups, and allows for partial credit if there are mistakes in the final answer or individual steps much like a standard mathematical examination.

    \item \textbf{LogicVista.} \cite{xiao_logicvista_2024} \textit{Accuracy.}
    Multiple-choice questions where reasoning skills are required.

    \item \textbf{MathVision.} \cite{wang_measuring_2024} \textit{Accuracy.}
    Open-ended and multiple choice math questions.

    \item \textbf{MMSafetyBench.} \cite{liu_mm-safetybench_2025} \textit{Attack Failure Rate.}
    Each open-ended question is intended to evaluate if a model can be manipulated to respond to unsafe queries, such as creating a computer virus. The \textit{Attack Failure Rate} we report is calculated as $100-\mathrm{Attack\ Success\ Rate}$ which is the original primary dataset metric, reflecting a successful attack. Using failure instead means that higher values are better in line with other dataset metrics allowing easy comparison.

    \item \textbf{CMMMU.} \cite{zhang_cmmmu_2024} \textit{Accuracy.}
    Inspired by MMMU, this Chinese dataset shares its objective and evaluation process. It includes multiple choice and open-ended questions.

    \item \textbf{WildVision.} \cite{lu_wildvision_2024} \textit{Win Rate.}
    Open-ended questions about images, with the response compared against a reference answer. The \textit{Win Rate} measures the fraction of responses better or much better than the reference.

    \item \textbf{MMBench CN.} \cite{liu_mmbench_2025} \textit{Accuracy (circular).}
    The Chinese version of MMBench, evaluated identically to MMBench EN.
\end{itemize}

\section{Approaches}
\label{sec:app:approaches}
Here we provide further details on the 34 different approaches (including parameter variations) we implement, with a full list shown in \cref{app:tab:approach_list}. Due to computational constraints, in line with prior work \cite{fu_deep_2026,zhu_towards_2026,xiaohu_know_2026} for all approaches that rely on the probability distribution at each token we use only the top-$k=20$ ones as an approximation to the complete distribution. All approaches are applied to each model independently.

Many approaches have seen prior use in the literature, but we introduce several new ones.

\paragraph{Mean Log Confidence:} Also referred to as average trace confidence or self-certainty, it uses the top-$k$ tokens at each position \cite{kang_scalable_2025,fu_deep_2026}:
\begin{equation}
C_{\log\text{avg}} = -\frac{1}{Nk}\sum_{i=1}^{N}\sum_{j=1}^{k}\log P_i(j),
\end{equation}
where $N$ is the sequence length, $k=20$ is the number of top tokens at each position $i$, and $P_i(j)$ is the probability assigned by the model to the $j$-th most likely token at position $i$.

Some prior work has combined the above calculation with Borda Voting \cite{kang_scalable_2025}, but as this is incompatible with open-ended questions we do not implement it.

\paragraph{Model Score (Normalized):} The Model Score is a natural result from the definition of sequential networks where we take into account the probability of each selected (by the sampler) token $\hat{j}$ in the sequence \cite{murray_correcting_2018,sutskever_sequence_2014}:
\begin{equation}
S_{\text{mod}} = \sum_{i=1}^{N} \log P_i(\hat{j}).
\end{equation}
The normalized variant divides the model score by the sequence length \cite{murray_correcting_2018,jean_montreal_2015}:
\begin{equation}
S_{\text{mod}}' = \frac{S_{\text{mod}}}{N}.
\end{equation}
Note that $S_{\text{mod}}'$ can be written in terms of the common language model metric perplexity (PPL) \cite{jelinek_perplexitymeasure_1977}: $S_{\text{mod}}' = -\log(\text{PPL}).$

\paragraph{Sequence Probability (Normalized):} The simplest possible measure based on the formulation of a sequential network \cite{sutskever_sequence_2014}, representing the probability of the final sequence:
\begin{equation}
    S_{\text{seq}} = \exp(S_{\text{mod}}) = \prod_{i=1}^{N} P_i(\hat{j}).
\end{equation}
We also evaluate a normalized variant (via the geometric mean) sometimes used as a baseline \cite{yao_reading_2025}:
\begin{equation}
    S_{\text{nseq}} = \exp(S_{\text{mod}}') = \left(\prod_{i=1}^{N} P_i(\hat{j})\right)^{\frac{1}{N}},
\end{equation}
which can also be tied to perplexity as the reciprocal \ie $S_{\text{nseq}} = \frac{1}{\text{PPL}}.$

\paragraph{Mean Confidence:} Similar to the \textit{Model Score}, but without the $\log$, it has been used as a baseline \cite{tan_basecal_2026}:
\begin{equation}
C_{\text{avg}} = \frac{1}{N} \sum_{i=1}^{N} P_i(\hat{j}).
\end{equation}

\paragraph{Mean/Maximum Entropy:} Entropy is a very well known method of measuring uncertainty, including with LLMs \cite{nguyen_probabilities_2026,kuhn_semantic_2023,fadeeva_fact-checking_2024}. Here, we take the Shannon Entropy of the top-$k$ tokens at each position, followed by the mean:
\begin{equation}
H_{avg} = -\frac{1}{N} \sum_{i=1}^{N} \sum_{j=1}^{k} P_i(j)\,\log P_i(j),
\end{equation}
and the maximum:
\begin{equation}
H_{\max} = \max_{1 \le i \le N} \left( -\sum_{j=1}^{k} P_i(j)\,\log P_i(j) \right).
\end{equation}

\paragraph{First/Last/Minimum Token:} Inspired by prior work that exploits very simple approaches \cite{chen_how_2026,zhao_first_2025,stacey_hidden_2026}, this involves using one of the tokens selected by the sampler as representative of the entire sequence:
\begin{equation}
\begin{aligned}
C_{\text{first}} &= P_1(\hat{j}),\\
C_{\text{last}} &= P_N(\hat{j}),\\
C_{\min} &= \min_{1 \le i \le N}\,  P_i(\hat{j}).
\end{aligned}
\end{equation}

\paragraph{Random:} This serves as a control, assigning a random score to each model from a uniform distribution.

\paragraph{Deep Confidence:} These approaches are based on \textit{Deep Think with Confidence} \cite{fu_deep_2026}, which we abbreviate by ``DC''. The original work focused on reasoning traces, but we use the entire token sequence including reasoning if applicable as not all evaluated models are capable of reasoning. We reproduce the key equations here for completeness. First, at the lowest level we have the \textit{token confidence} $C_i$, defined at every position $i$ and using the top-$k$ tokens:
\begin{equation}
    C_{i} = -\frac{1}{k} \sum_{j=1}^{k} \log P_{i}(j).
\end{equation}
We can then calculate \textit{group confidence}, where we have a sliding window group $G_i$ composed of $n$ consecutive tokens with overlapping adjacent windows:
\begin{equation}
    \label{eq:app:group_conf}
    C_{G_{i}} = \frac{1}{|G_{i}|} \sum_{t \in G_{i}} C_{t},
\end{equation}
where $|G_{i}|$ is the number of tokens in the group. We can now define the evaluated metrics, beginning with the ``Bottom $\alpha\%$'' variants:
\begin{equation}
    C_{\text{bottom-}\alpha\%} = \frac{1}{|G_{b}|} \sum_{G_{j} \in G_{b}} C_{G_{j}},
\end{equation}
where $G_b$ is the set of groups with the lowest bottom-$\alpha\%$ scores. Next, we have the minimum group confidence, where we use the least confident group as a proxy for the entire sequence:
\begin{equation}
    C_{\min G} = \min_{G_{j} \in G} C_{G_{j}},
\end{equation}
where $G$ is the set of all groups. Lastly, we have the ``Tail'' metrics:
\begin{equation}
    C_{\text{tail}} = \frac{1}{|T_{\text{tail}}|} \sum_{t \in T_{\text{tail}}} C_{t},
\end{equation}
where $T_{\text{tail}}$ includes either a fixed number or a percentage of tokens from the end of the sequence.

We note that as shown in \cref{app:tab:approach_list}, many ``DC'' approaches include an integer number such as 8,16, or 2048: with the exception of the ``Tail'' variants, this specifies the size of the sliding window used in \cref{eq:app:group_conf}.

\paragraph{P(Confidence):} Prior work has shown that some tokens in a model's response are more important than others and that models can extract the most important token from their own answer \cite{duan_shifting_2024,orgad_llms_2025}. To the best of our knowledge, however, there is no work that combines the two into a unified confidence estimation approach relying only on the model's response without a probe of the model. As such, we create a unified approach by asking the model to find the ``most important'' string in its own prior response via the following prompt:

\begin{promptbox} 
From your previous answer, select the shortest contiguous substring that best represents your certainty in the answer’s core claim.\\
Guidelines:
\begin{itemize}[nosep,leftmargin=*]
    \item Prefer the main asserted content (the key claim, conclusion, or decisive phrase).
    \item If an error was made in a reasoning step, prefer the string containing the error.
    \item Avoid purely grammatical or filler words (\eg, ``the'', ``a'', ``there is'').
    \item The substring must appear verbatim in the previous answer.
    \item Do not paraphrase, add, or remove words.
    \item If the answer contains multiple sentences, select the most decisive part.
    \item If no clear confidence-relevant phrase exists, return the entire answer.
\end{itemize}
Output ONLY the substring.
\end{promptbox}
We then search for that string in the model's \emph{original} response and use the geometric mean of the probability of each matched token:
\begin{equation}
    \label{eq:app:p_conf}
    C_{P(\text{conf})} = \left(\prod_{t=a}^{b} P_t(\hat{j})\right)^{\frac{1}{b-a+1}},
\end{equation}
where $a$ and $b$ indicate the start and end positions of the matched string. We note that many older models in particular struggle to follow the prompt's instructions, which make matching difficult.

\paragraph{P(True):} One of the first to explore a model assessing its own response \cite{kadavath_language_2022}; including variants it remains a popular baseline \cite{farquhar_detecting_2024,orgad_llms_2025,agnimo_evaluating_2026,tan_basecal_2026,cao_clustered_2026}. We evaluate several versions of P(True), including some of which we introduce, all of which start with the same prompt given to the model after its response to a question:
\begin{promptbox} 
Determine whether your immediately previous answer was factually correct for the given query.\\
Be critical of your answer; it may be incorrect.\\
If the previous answer was correct, output exactly:\\
True\\
If the previous answer was incorrect, output exactly:\\
False\\
Output exactly one of these two words. Do not include any other text, punctuation, or explanation.
\end{promptbox}
We refer to the model's response to this prompt as the P(True) response, differentiating it from the original response to the question. We can now implement the basic P(True) approach, following the basic process although not the specific implementation of \cite{kadavath_language_2022}, where we match on the model's P(True) response containing keywords indicating correctness or failure:
\begin{equation}
\label{eq:app:conf_p_true}
C_{P(\text{True})} =
\begin{cases}
1, & r = \texttt{true}, \\
0, & r = \texttt{false}, \\
0, & \exists\, u \in \mathcal{U}:\; u \preceq r, \\
1, & \exists\, s \in \mathcal{T}_{+}:\; s \prec_w r \;\land\;
     \not\exists\, s \in \mathcal{T}_{-}:\; s \prec_w r, \\
0, & \exists\, s \in \mathcal{T}_{-}:\; s \prec_w r \;\land\;
     \not\exists\, s \in \mathcal{T}_{+}:\; s \prec_w r, \\
0, & \text{otherwise},
\end{cases}
\end{equation}
where $\mathcal{U}$ is a set of common uncertainty phrases (\eg
``not sure'', ``I don't know''), $r$ denotes the P(True) response, $s \prec_w r$ denotes that keyword $s$ occurs in $r$ as a complete word, and $u \preceq r$ means that phrase $u$ occurs in $r$. We also strip some formatting and perform all matches without regard to case. The keywords in each set are:
\begin{equation}
\begin{aligned}
\mathcal{T}_{+} &= \{\,
  \texttt{true},\;\texttt{yes},\;\texttt{correct},\\
  &\qquad
  \texttt{accurate},\;\texttt{right},\;\texttt{affirmative}\,\},\\
\mathcal{T}_{-} &= \{\,
  \texttt{false},\;\texttt{no},\;\texttt{incorrect},\\
  &\qquad
  \texttt{inaccurate},\;\texttt{wrong},\;\texttt{uncertain},\\
  &\qquad
  \texttt{unsure}\,\}.
\end{aligned}
\end{equation}

\paragraph{P(True) Confidence (Normalized):} This version is based on \cite{xiaohu_know_2026}, where we examine only the first token of the P(True) response and match on the ``True'' token in the top-$k=20$ tokens. The probability of this token is then used to represent the model's confidence in the original answer. This assumes that ``True'' is a single token in the tokenizer, which is the case for any standard model.
\begin{equation}
    \label{eq:app:p_true_exp_conf}
    C_{\text{P(True) C.}} = P_1(j=\text{True}).
\end{equation}
We also implement a normalized version where we implement the normalization process of \cite{xiaohu_know_2026}. This involves searching for the ``True'' and ``False'' tokens in the top-$k$ tokens at the first position and calculating:
\begin{equation}
    \label{eq:app:p_true_exp_conf_norm}
    C_{\text{P(True) C. Norm.}} = \frac{P_1(j=\text{True})}{P_1(j=\text{True}) + P_1(j=\text{False})}.
\end{equation}
\paragraph{P(True) Search (Normalized):} \textit{P(True) Confidence} is unable to properly handle models where the first token in the P(True) response is something other than ``True'' or ``False''. Some models start with formatting (\eg **), while reasoning models start with a special thinking token (\eg \textless think\textgreater). We introduce a simple heuristic to work around this: starting from the end of the P(True) response, search for the first token in the response that is a case-insensitive match for ``true'' or ``false'', and then apply \cref{eq:app:p_true_exp_conf} at the matching token position. Thus we have:
\begin{equation}
    \label{eq:app:p_true_srch}
    \begin{aligned}
            i &= \search_{N \ge i \ge 1}(\{\texttt{true} \lor \texttt{false}\})\\
            C_{\text{P(True) Srch.}} &= P_i(j \simeq \text{True}),
    \end{aligned}
\end{equation}
where $\simeq$ represents case insensitive matching.
For the normalized variant, the search process is the same but we use \cref{eq:app:p_true_exp_conf_norm} and further expand the matching criteria:
\begin{equation}
    \label{eq:app:p_true_srch_norm}
    C_{\text{P(True) Srch. Norm.}} = \frac{\sum_{j \approx \text{True}} P_i(j)}{\sum_{j \approx \text{True}}P_i(j) + \sum_{j \approx \text{False}}P_i(j)},
\end{equation}
where $\approx$ allows for a match if a token contains the substring ``true'' or ``false''.

\paragraph{P(True) All Tokens:} While \textit{P(True) Search} finds a single True/False token, the model's response often includes the tokens True and False somewhere within the top-$k=20$ tokens at each position. This approach uses all of them:
\begin{equation}
    \label{eq:app:p_true_all_token}
    C_{\text{P(True) All Tok.}} =
    \frac{1}{|\mathcal{S}|} \sum_{i \, \in \, \mathcal{S}}
    \frac{
    \sum_{j \approx \text{True}} P_i(j)}{
    \sum_{j \approx \text{True}} P_i(j) +
    \sum_{j \approx \text{False}} P_i(j)},
\end{equation}
where $\mathcal{S}$ is the set of all tokens $i \in \{1, \dots, N\}$ where at least one of the top-$k$ tokens at that position contains ``true'' or ``false''.

\paragraph{P(True) Entropy:} This final variant is inspired by entropy-based knowledge distillation \cite{zhang_egad_2026}, and is intended to be a parameter-free approach that scales \textit{P(True) Search Normalized} based on the model's output distribution. This is particularly relevant when using diverse models, where one model may assign very high confidence to all tokens in a sequence but another may be more measured.
We begin by computing the mean normalized entropy across the original response:
\begin{equation}
    \label{eq:app:p_true_entropy_scale}
    \bar{h} = \frac{1}{N} \sum_{i=1}^{N} \frac{H(\mathbf{P}_i)}{\log k},
\end{equation}
where $\mathbf{P}_i = (P_i(1),\ldots,P_i(k))$ is the top-$k=20$ token probability vector at position $i$ of the \emph{original} response, $N$ is the total number of tokens in that response, and $H(\mathbf{P}_i)$ is the Shannon entropy of the distribution at that position.
We then apply the scaling:
\begin{equation}
    \label{eq:app:p_true_entropy}
      C_{\text{P(True) Entropy}} = \log \frac{C_{\text{P(True) Srch. Norm.}}}{1 - C_{\text{P(True) Srch. Norm.}}} \cdot \bar{h},
\end{equation}
which centers the confidence score around $0$.

Lastly, we note that for \cref{eq:app:p_true_all_token,eq:app:p_true_srch_norm,eq:app:p_true_entropy,eq:app:p_true_srch} in edge cases where one of ``True'' or ``False'' is missing in the top-$k$ tokens, but the other is defined, we use $P_i(j \approx \text{True})$ as is or $1-P_i(j \approx \text{False})$ as necessary.

\begin{table}
\centering
\caption{List of all the approaches tested in this work. The \textit{All Plots} column indicates if the approach is relevant enough to be shown on all plots, as due to space constraints we elected to only use a subset in most plots with some exceptions such as \cref{fig:app:relax_win_rate_appr_ensemble_heatmap}. Abbreviations: L. - Length, Norm. - Normalized, Prob. - Probability, Tok. - Tokens, Srch. - Search, Conf. - Confidence, Min. - Minimum, Max. - Maximum, Bott. - Bottom, W. - Width.}
\label{app:tab:approach_list}
    \begin{tabular}{lc}
    \toprule
    Approach & All Plots\\
    \midrule
    Model Score & \cmark \\
    Model Score (L. Norm.) & \cmark \\
    Mean Log Confidence & \cmark \\
    Mean Confidence & \cmark \\
    Mean Entropy & \cmark \\
    Sequence Prob. & \cmark \\
    Sequence Prob. (Norm.) & \cmark \\
    P(Confidence) & \cmark \\
    P(True) & \cmark \\
    P(True) All Tok. & \xmark \\
    P(True) Srch. (Norm.) & \cmark \\
    P(True) Conf. (Norm.) & \cmark \\
    P(True) Entropy & \cmark \\
    Random & \cmark \\
    P(True) Conf. & \cmark \\
    P(True) Srch. & \cmark \\
    Max. Entropy & \cmark \\
    1st Token Conf. & \cmark \\
    Last Token Conf. & \cmark \\
    Min. Conf. & \cmark \\
    DC Tail 10\% & \cmark \\
    DC Tail 20\% & \xmark \\
    DC Tail 8 & \xmark \\
    DC Tail 16 & \cmark \\
    DC Tail 2048 & \cmark \\
    DC Min. 8 & \cmark \\
    DC Min. 16 & \xmark \\
    DC Min. 2048 & \cmark \\
    DC Bott. 10\%, W. 8 & \xmark \\
    DC Bott. 10\%, W. 16 & \xmark \\
    DC Bott. 10\%, W. 2048 & \xmark \\
    DC Bott. 50\%, W. 8 & \xmark \\
    DC Bott. 50\%, W. 16 & \xmark \\
    DC Bott. 50\%, W. 2048 & \xmark \\
    \bottomrule
    \end{tabular}
\end{table}

\section{Ensemble Configuration}
\label{sec:app:ensemble_config}

In \cref{tab:app:ensemble_groups}, we list all the ensembles we evaluate and their constituent models. \bestofn approaches work with only a single model, while other more diverse approaches combine results from different models which may not share an architecture. The constituent models are chosen at the time of evaluation and writing to be recent open-weight architectures known to generally perform well on LVLM tasks but small enough to run on a local machine.

For the \textit{Refine and Align} and \textit{Vision Verification} ensembles, we included older models as specified in the publications but were unable to evaluate the full set used in each \cite{tekin_vision_2026,pandey_refine_2025}. We encountered issues with some models such as a mismatch between the model definition and provided weights, compatibility issues with VLLM, or context sizes too small to handle most datasets we use. Thus, our choice of models is restricted to those compatible with VLLM and capable of processing the majority of the datasets we evaluate on.

\begin{table*}[t]
    \centering
    \caption{Complete list of ensemble configurations used in our experiments and their constituent models. See \cref{tab:app:individual_mod_results,tab:app:sampling_params} for the performance of each model, relevant citations, and sampling parameters.}
    \label{tab:app:ensemble_groups}

    \renewcommand{\arraystretch}{1.15}

    \begin{tabularx}{\textwidth}{
        >{\raggedright\arraybackslash}p{0.27\textwidth}
        >{\raggedright\arraybackslash}X
    }
        \toprule
        \textbf{Description} & \textbf{Models} \\
        \midrule

        \multicolumn{2}{l}{\textit{\bestofn Approaches}} \\
        \midrule

        Qwen 3.5 \bestofn Instruct
        & \model{Qwen 3.5 9B Instruct} \\

        Qwen 3.5 \bestofn Thinking
        & \model{Qwen 3.5 9B Thinking} \\

        GLM 4.6V \bestofn Instruct
        & \model{GLM 4.6V Flash Instruct} \\

        GLM 4.6V \bestofn Thinking
        & \model{GLM 4.6V Flash Thinking} \\

        \midrule
        \multicolumn{2}{l}{\textit{Diverse Approaches}} \\
        \midrule

        Refine and Align
        & \model{LLaVA-OneVision Qwen2-OV 7B Instruct},
          \model{Gemma 3 4B Instruct},
          \model{Qwen 2.5-VL 3B Instruct} \\

        Vision Verification
        & \model{LLaVA v1.6 Vicuna 7B Instruct},
          \model{LLaVA v1.6 Vicuna 13B Instruct},
          \model{Qwen 2.5-VL 7B Instruct},
          \model{InternVL 2 8B Instruct} \\

        Modern 1
        & \model{GLM 4.6V Flash Instruct},
          \model{Qwen 3-VL 8B Instruct} \\

        Modern 2
        & \model{GLM 4.6V Flash Instruct},
          \model{Qwen 3.5 9B Instruct} \\

        Modern 3
        & \model{GLM 4.6V Flash Instruct},
          \model{Qwen 3.5 9B Instruct},
          \model{InternVL 3.5 8B Instruct} \\

        Modern 4
        & \model{GLM 4.6V Flash Instruct},
          \model{Qwen 3.5 9B Instruct},
          \model{InternVL 3.5 8B Instruct},
          \model{Ministral 3 8B 2512 Instruct} \\

        Small Models
        & \model{SmolVLM2 2B Instruct},
          \model{Qwen 3.5 2B Instruct},
          \model{Ovis 2.5 2B Instruct},
          \model{InternVL 3.5 2B Instruct} \\

        Different Size 1
        & \model{Qwen 3.5 2B Instruct},
          \model{Qwen 3.5 4B Instruct},
          \model{Qwen 3.5 9B Instruct} \\

        Different Size 2
        & \model{InternVL 3.5 2B Instruct},
          \model{InternVL 3.5 8B Instruct} \\

        Old \& New 1
        & \model{MiniCPM V4 Instruct},
          \model{MiniCPM V4.5 Instruct} \\

        Old \& New 2
        & \model{Qwen 3-VL 8B Instruct},
          \model{Qwen 3.5 9B Instruct} \\

        Old \& New 3
        & \model{MiniCPM V4 Instruct},
          \model{MiniCPM V4.5 Instruct},
          \model{Qwen 3-VL 8B Instruct},
          \model{Qwen 3.5 9B Instruct} \\

        Architecture Variation 1
        & \model{InternVL 3.5 8B Instruct},
          \model{MiniCPM V4.5 Instruct},
          \model{Kimi-VL A3B Instruct},
          \model{Ministral 3 8B 2512 Instruct} \\

        Architecture Variation 2
        & \model{Qwen 3.5 9B Instruct},
          \model{Ovis 2.5 9B Instruct},
          \model{Ministral 3 8B 2512 Instruct} \\

        Architecture Variation 3
        & \model{Qwen 3.5 4B Instruct},
          \model{Gemma 3 4B Instruct} \\

        Similar Architecture Variation
        & \model{Ovis 2.5 9B Instruct},
          \model{Qwen 3-VL 8B Instruct},
          \model{MiniCPM V4.5 Instruct},
          \model{InternVL 3.5 8B Instruct} \\

        Instruct vs.\ Thinking: Instruct
        & \model{GLM 4.6V Flash Instruct},
          \model{MiniCPM V4.5 Instruct},
          \model{Ministral 3 8B 2512 Instruct},
          \model{Qwen 3.5 9B Instruct} \\

        Instruct vs.\ Thinking: Thinking
        & \model{GLM 4.6V Flash Thinking},
          \model{MiniCPM V4.5 Thinking},
          \model{Ministral 3 8B 2512 Thinking},
          \model{Qwen 3.5 9B Thinking} \\

        Instruct \& Thinking
        & \model{GLM 4.6V Flash Instruct},
          \model{Qwen 3.5 9B Thinking} \\

        Top Thinking Performers
        & \model{Step 3-VL 10B Thinking},
          \model{Qwen 3.5 9B Thinking},
          \model{MiniCPM V4.5 Thinking} \\

        \bottomrule
    \end{tabularx}
\end{table*}

\section{Comparison Issues}
\label{sec:app:comparison_issues}
When creating \cref{tab:res:refine_verfication_compare}, we attempted to recreate the exact ensembles used in the original work, but were unable to duplicate them exactly due to compatibility issues with certain models, including mismatches between provided model weights and the architecture. We are also aware that there is a bit of a mismatch with the datasets and metrics used; for instance, the VQARad dataset \cite{lau_dataset_2018} used by \cite{pandey_refine_2025} is specific to medicine and thus we do not evaluate it in our work. On the other hand, we attempted to evaluate the MMMU Pro dataset \cite{yue_mmmu-pro_2025} as used by \cite{tekin_vision_2026} but observed that many questions were not compatible (\eg exceeding context) with several of the older models. Nonetheless, we feel that we can still make a fair comparison. This is substantiated by the fact that unrelated approaches \cite{zhang_why_2026, wang_seeing_2026,bugaud_hidden_2026} using different LVLMs on various datasets report relative gain results roughly in line with \cref{tab:res:refine_verfication_compare}, with the worst being $-1.58\%$ and many in the single digit range \eg $+1.24\%$, $+2.88\%$ or $+5.14\%$. Some of the best results exceed the relative performance of the approaches we test (\eg $+15.96\%$), but this is expected given the extensive task-specific scaffolding used.

\section{On Calibration}
In this work, we evaluate different confidence measures for model selection. Prior work has shown that for individual models calibration error and accuracy are not correlated \cite{plaut_probabilities_2025}, which leaves only the relative calibration between models; this can be useful especially when studying ensembles of different model architectures. We do not take this relative calibration into account, as we are evaluating means of self-assessment, and it is unclear a priori how calibration may affect any given approach: some may need it to function well, others may not. This allows us to (ideally) identify an approach that can perform well under a wide variety of circumstances without a dependency on a calibration process. Furthermore, there are many potential variables involved in calibration (choice of holdout training set if applicable, calibration algorithm) with some recent promising approaches \cite{tan_basecal_2026} requiring model variants that are not available for many of the models we test. These variables add significant complexity to the experiment process, and thus we leave them for future work to be studied separately.

\section{Approach Comparison}
\label{sec:app:approach_compare}
\begin{figure}[ht]
    \centering
    \includegraphics[width=\linewidth]{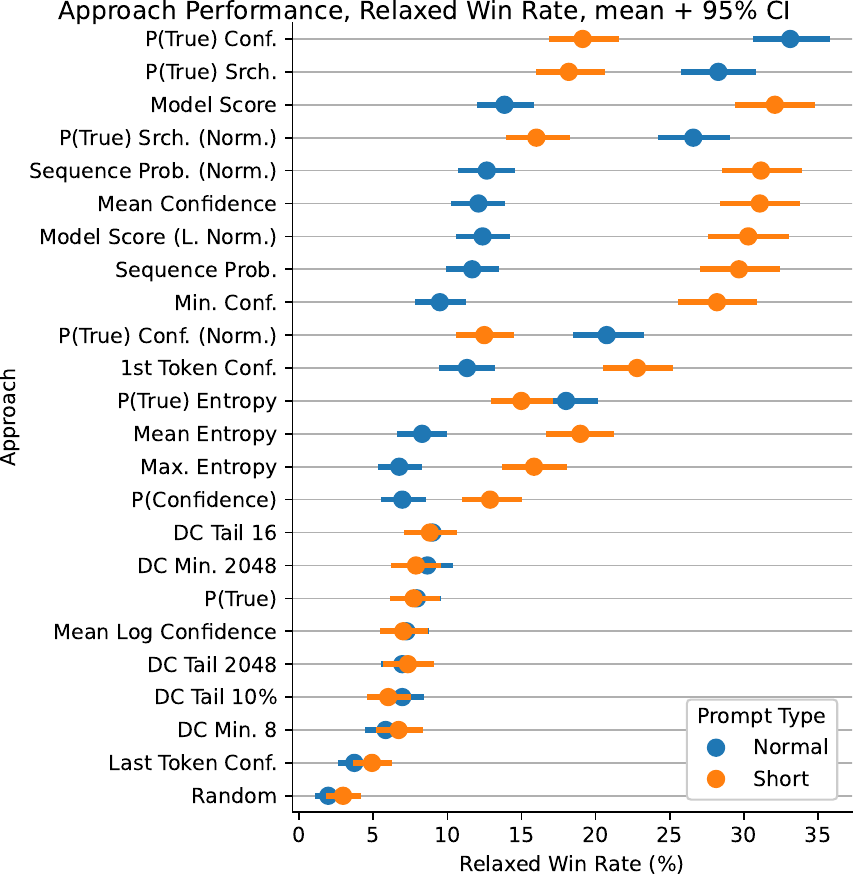}
    \caption{Comparison of different model selection approaches, presented as the relaxed average win rate (within $0.5\%$ of the winner) across all datasets, ensemble configurations, prompt types and seeds. Results show the mean and $95\%$ confidence interval. The confidence interval is calculated using bootstrapping with 2000 iterations.}
    \label{fig:app:relax_win_rate_plot}
\end{figure}

In \cref{fig:app:relax_win_rate_plot}, we compare approaches by the \textit{Relaxed Win Rate}, which we define as the proportion of experiments (\%) where, for an experiment run, an approach achieved the best performance or was within $0.5\%$.

In \cref{fig:app:relax_win_rate_appr_ensemble_heatmap}, we use the same data as \cref{fig:app:relax_win_rate_plot}, but drop the prompt type variable in favour of showing the impact of the ensemble configuration. We see that our choice to exclude certain approaches in the interest of brevity such as ``DC Bott. 50\%'' is justified, as the performance is not substantially different from similar included approaches. The main takeaway is that there is no clear winner with the choice of approach being dictated by many variables: \cref{fig:app:relax_win_rate_plot} shows that prompt changes have a significant effect, and \cref{fig:app:relax_win_rate_appr_ensemble_heatmap} shows that the ensemble also has a significant effect.

\begin{figure}[ht]
    \centering
    \includegraphics[width=\linewidth]{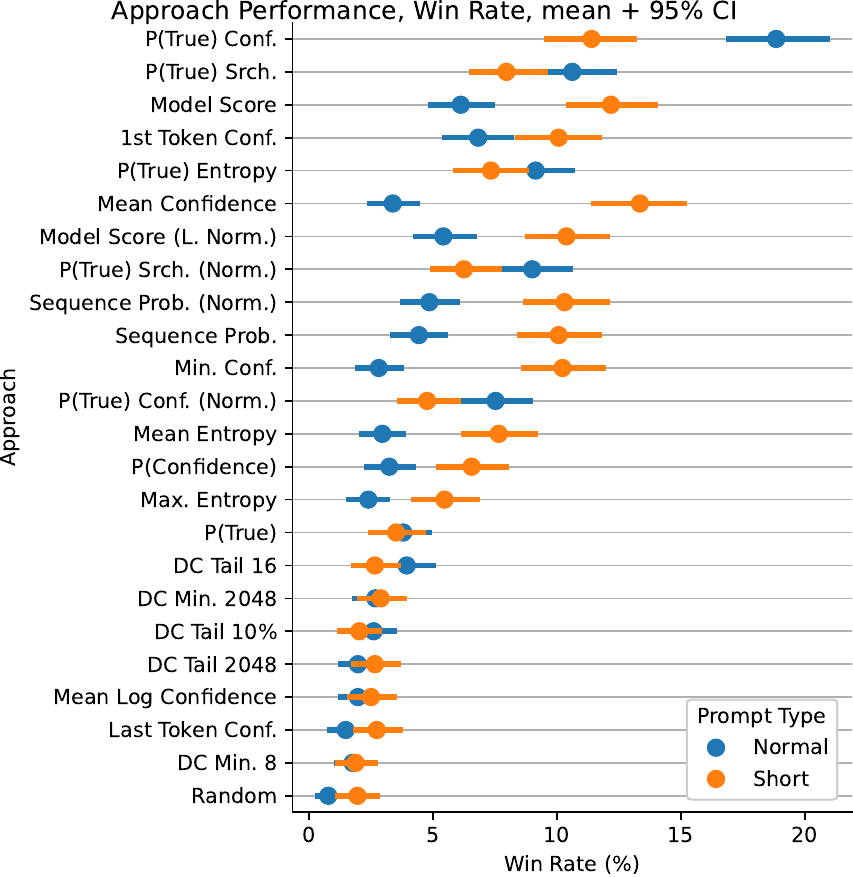}
    \caption{Comparison of different model selection approaches, presented as the average win rate across all datasets, ensemble configurations, prompt types and seeds. Results show the mean and $95\%$ confidence interval. The confidence interval is calculated using bootstrapping with 2000 iterations. This is the same plot as \cref{fig:app:relax_win_rate_plot} except we use a strict win rate definition.}
    \label{fig:app:strict_win_rate_plot}
\end{figure}

\begin{figure*}[p]
    \centering
    \includegraphics[width=\linewidth]{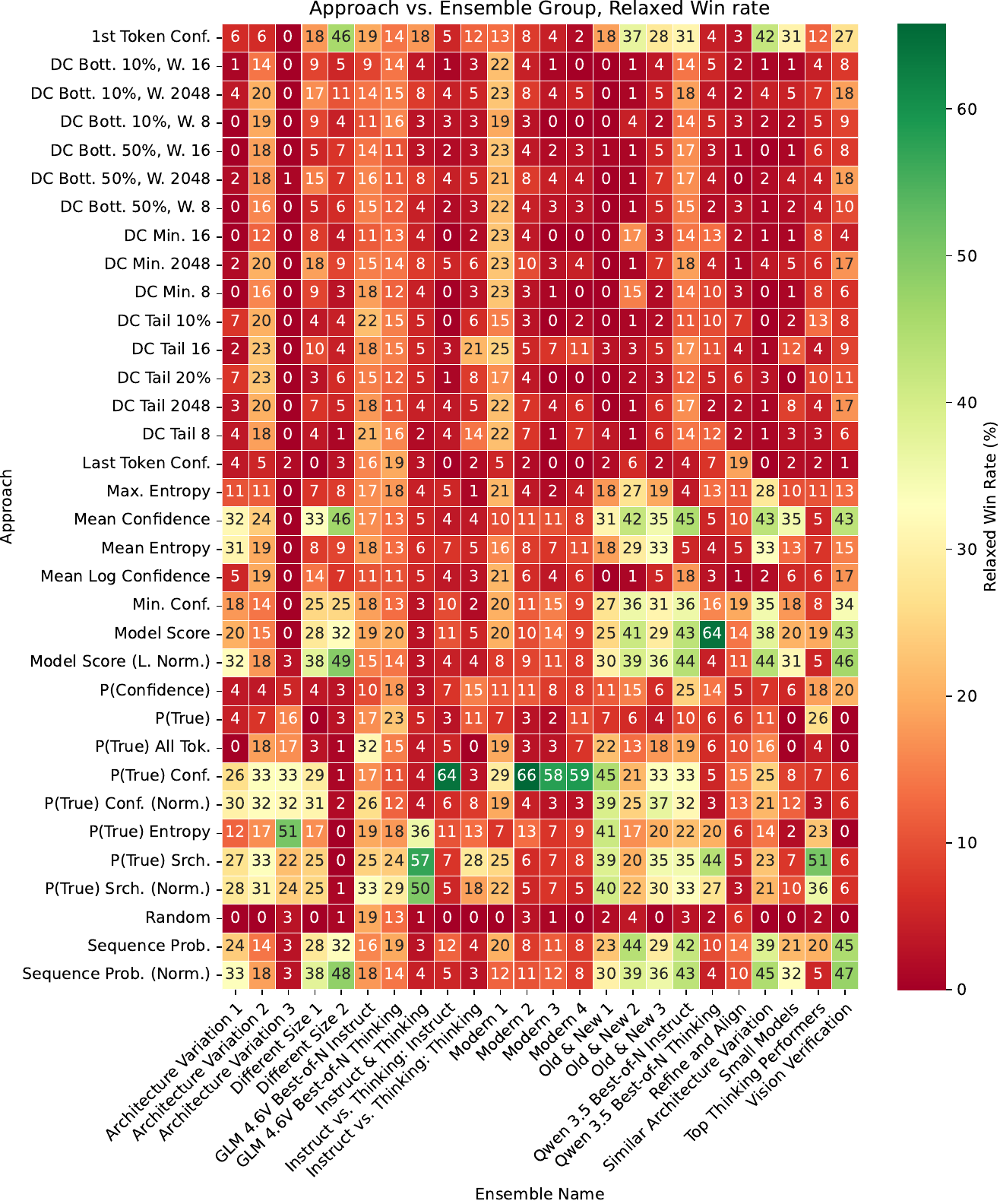}
    \caption{Comparison of different model selection approaches against ensemble configurations. The average relaxed win rate (within $0.5\%$ of the winner) is shown across prompt types, datasets, and seeds. Unlike \cref{fig:app:relax_win_rate_plot,fig:app:strict_win_rate_plot} we include the full set of approaches.}
    \label{fig:app:relax_win_rate_appr_ensemble_heatmap}
\end{figure*}

\cref{fig:app:strict_win_rate_plot} is a further variant of \cref{fig:app:relax_win_rate_plot}, except that we employ a strict interpretation of the win rate. Thus, even an approach that is extremely close in performance to the winner is considered a loss. However, approaches that exactly match the winner performance are included, for the few rare cases where this occurs. We include this comparison to show that relaxing the win rate avoids excessive punishment of approaches that still perform very well, but even with a strict interpretation our conclusions still hold.

\section{Additional Findings}
\label{sec:app:additional_findings}

\paragraph{Dataset category has minimal impact:} If we zoom in on the dataset variable in \cref{fig:res:heatmap_percent_gain_dataset_best}, we note that there does not appear to be any clear correlation between performance and the category the dataset belongs to. Specifically, we do not see any correlation between the task or category of a given dataset (\eg OCR or math problems) and the maximum achievable performance for any given ensemble configuration.

\paragraph{There is no one best approach:}  In \cref{fig:app:relax_win_rate_plot}, the best approaches only manage about 30\%, with many of the leaders such as P(True) variants or the Model Score being clustered together. Of note is that while P(True) Conf. is in the lead, this is something of a spurious result, as it is caused by P(True) Conf. struggling to handle the response format of the GLM models in several ensemble configurations \eg the ``Modern'' set. This leads to other constituent models which perform better and do not suffer from the same failure mode (\eg Qwen, see \cref{tab:app:individual_mod_results}) being used. While one could argue that this is intended behaviour, where a model that does not strictly follow instructions is rejected leading to higher performance, we consider P(True) Search to be the overall leader. Please see \cref{sec:app:p_true_spurious} for further details.

\paragraph{High Sensitivity to small details:} In \cref{fig:app:relax_win_rate_plot}, we see that P(True) variants, where we use the token-level confidence of the model's attempt to predict the correctness of its answer, are in the lead, with various approaches that aggregate the token-level confidence in the model's answer to the question trailing behind. However, the basic P(True) approach, which relies on the response containing the string ``True'' or ``False'', does not perform very well, in line with \cite{fadeeva_don_2026,srey_learning_2026}. Please see \cref{sec:app:p_true_spurious} for a specific example on how impactful one model's particular quirk (a small detail) can be. More generally, very simple approaches such as relying on the probability of the first token also work surprisingly well, while more complex approaches such as \textit{Deep Confidence } \cite{fu_deep_2026} or P(Confidence) \cite{duan_shifting_2024,orgad_llms_2025} encounter difficulty generalizing beyond their original use cases. This high degree of sensitivity to a range of variables, including what could be considered minor details such as how token-level information is processed, makes it difficult for any one approach to generalize.

\paragraph{Normalization benefits unclear:} Comparing approaches to their normalized versions in \cref{fig:app:relax_win_rate_plot}, we see that normalization is not always useful. This is particularly true for P(True), where in contradiction to \cite{xiaohu_know_2026} across the many datasets and models normalization is detrimental.

\paragraph{Absolute performance cannot be ignored:} In \cref{tab:app:individual_mod_results}, we show the individual performance of each model, while \cref{tab:app:abs_ensemble_vs_approach_common_ds,tab:app:abs_ensemble_vs_approach_common_ensemble,tab:app:abs_ensemble_vs_dataset_best_approach,tab:app:abs_ensemble_vs_dataset_ideal_approach} present the performance of ensembles along several axes, such as dataset or approach. While we emphasize relative performance as a means of assessing ensemble effectiveness, absolute performance remains important; there are many ensemble configurations where relative performance may exceed the constituent models, but better absolute performance can be achieved with a different ensemble or even an individual model.

The prompt variable is noteworthy for emphasizing the role of absolute performance. If we examine \cref{tab:app:abs_ensemble_vs_approach_common_ensemble} and use ``Architecture Variation 2'' as an example, we see the \textit{Best Model} score is $75.61$ for the ``Normal'' prompt, but only $65.74$ for the ``Short'' prompt as some constituent models perform worse when asked to be concise. Meanwhile, token-level ensembling approaches such as ``Mean Log Confidence'' struggle with longer responses and thus deliver poor relative performance but better absolute performance ($74.53$) compared to the shorter prompt ($66.25$) case.

\section{Individual Model Results}
\label{sec:app:individual_results}
In \cref{tab:app:individual_mod_results}, we report the results for each individual model on each dataset. This is a reference useful to compare ensemble performance against constituent models. Note that two edge cases appear in the data. First, some combinations are impossible to run due to compatibility issues, described in \cref{sec:app:dataset_details} and denoted by ``-'' in the cell with a grey background. Second, in cases where the model was able to run, the performance is reported to be approximately 0. Such poor performance is caused by a mismatch between the model output and the dataset evaluation. We observe such cases to occur when the dataset evaluation relies on heuristics, such as string matching, and the model output violates the underlying formatting assumptions despite being correct in many cases. An example of this is OCRVQA and the GLM 4.6V model, where the model often outputs a variant of ``The answer is X'' and string matching is unable to handle the short sentence even though ``X'' is correct. A potential solution for such cases would be to use LLM-as-a-judge instead of the default string matching. As at least some models are able to handle these cases without issue, however, we consider such targeted per-dataset intervention beyond our scope. It furthermore serves as a difficult test case during evaluation, as exposes gaps in the various self-assessment approaches we test when ensembling.

\section{Ensemble Group Comparison}
\label{sec:app:more_ensemble_compare}
\begin{figure*}
    \centering
    \includegraphics[width=\linewidth]{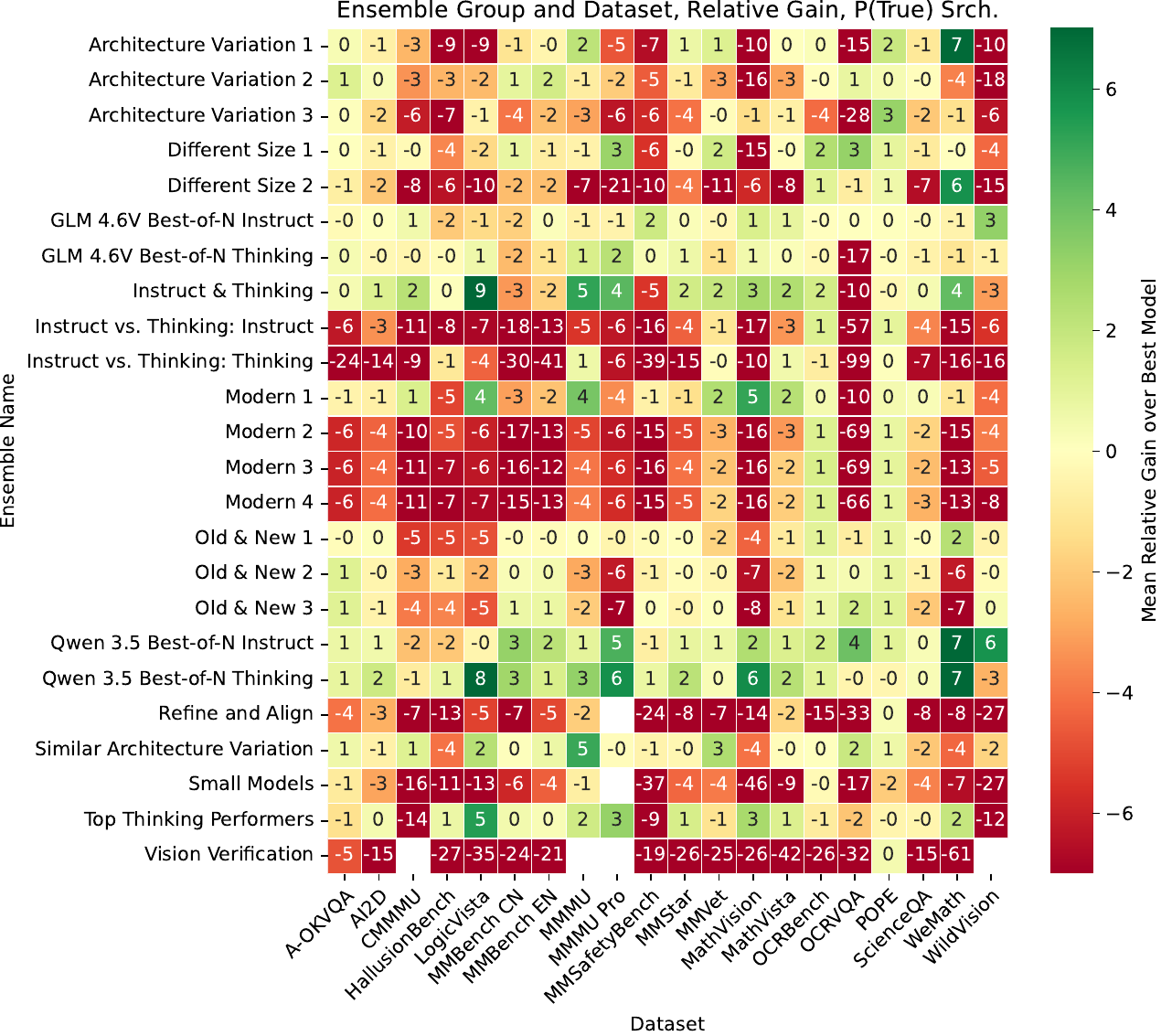}
    \caption{Comparison of different ensembles by dataset, presented as the \textbf{Relative Gain (\%)} over the best possible performance using one of the models that make up the ensemble, chosen optimally for each dataset, prompt and seed combination. This is the same as \cref{fig:res:heatmap_percent_gain_dataset_best} but with the approach fixed to P(True) Search instead of the best possible.}
    \label{fig:app:heatmap_percent_gain_dataset_p_true_search}
\end{figure*}

\begin{figure*}
    \centering
    \includegraphics[width=\linewidth]{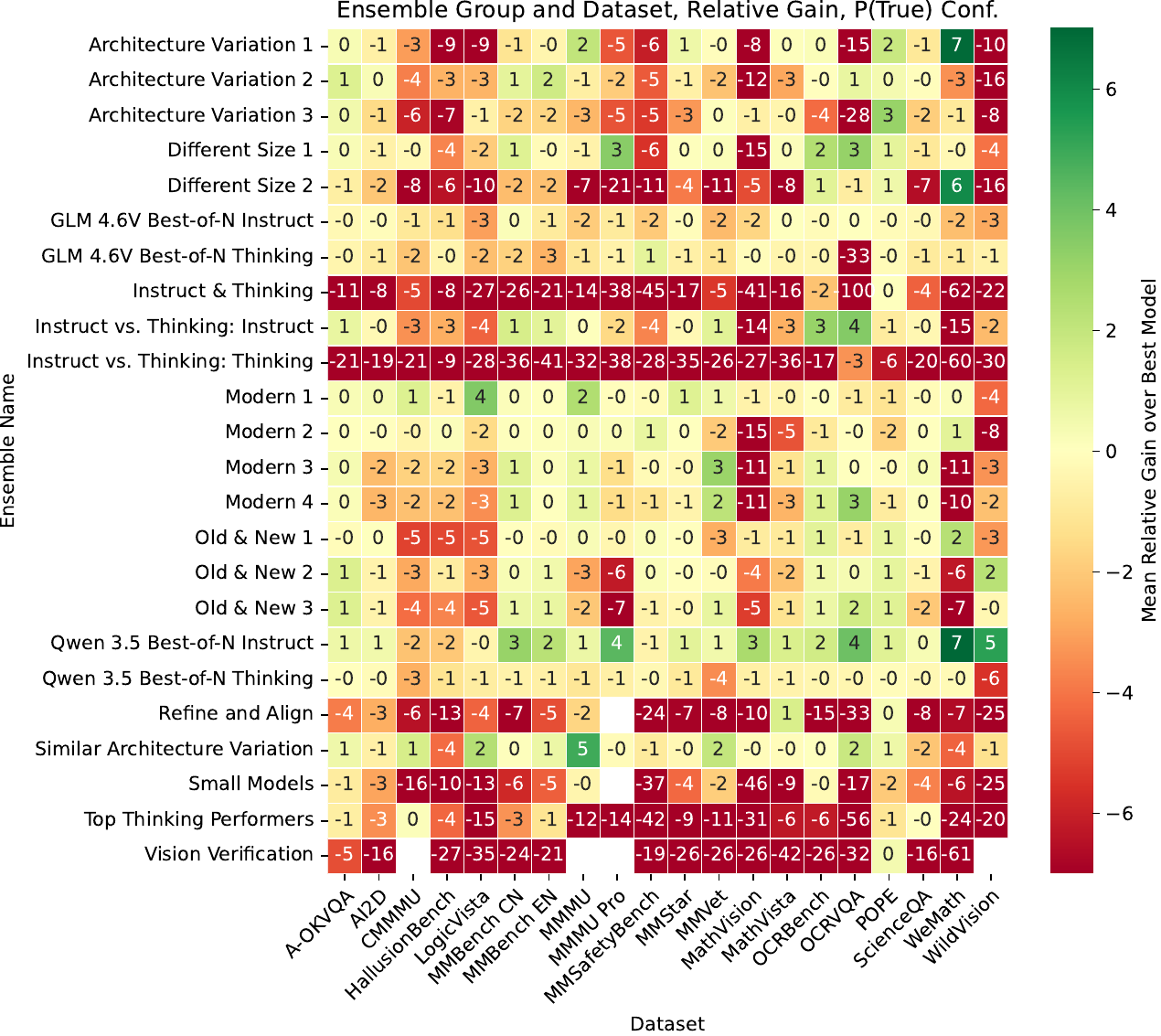}
    \caption{Comparison of different ensembles by dataset, presented as the \textbf{Relative Gain (\%)} over the best possible performance using one of the models that make up the ensemble, chosen optimally for each dataset, prompt and seed combination. This is the same as \cref{fig:res:heatmap_percent_gain_dataset_best} but with the approach fixed to P(True) Confidence instead of the best possible.}
    \label{fig:app:heatmap_percent_gain_dataset_p_true_conf}
\end{figure*}

In \cref{fig:app:heatmap_percent_gain_dataset_p_true_search,fig:app:heatmap_percent_gain_dataset_p_true_conf}, we present the same heatmap as \cref{fig:res:heatmap_percent_gain_dataset_best} but fix the approach to P(True) Search and P(True) Confidence respectively instead of using the best possible one. The difference is quite stark, with many combinations showing double-digit losses. Nonetheless, it remains quite possible to find, for a given ensemble, several datasets with reasonable performance.

We also provide several large and detailed tables comparing the performance of various ensemble, approach and dataset combinations. Each cell shows the absolute numerical performance, with cell shading indicating relative performance. All important conclusions are already covered in the main paper in an easier to analyze format, but we include these tables for completeness to enable the interested reader to drill down into the details of how the different variables in play interact. 

In \cref{tab:app:abs_ensemble_vs_approach_common_ds}, we compare the performance of various ensembles against the approach used, across the 15 datasets all are compatible with. In \cref{tab:app:abs_ensemble_vs_approach_common_ensemble}, we compare the performance of various ensembles against the approach used, but unlike \cref{tab:app:abs_ensemble_vs_approach_common_ds} we use all datasets and do not show the 3 ensemble configurations that were unable to be run against all datasets. In \cref{tab:app:abs_ensemble_vs_dataset_best_approach}, we compare each ensemble's performance on a per-dataset basis. The \textit{P(True) Search} approach is used here, being the best one we identified. In \cref{tab:app:abs_ensemble_vs_dataset_ideal_approach}, we compare each ensemble's performance on a per-dataset basis using the best possible approach for every ensemble, prompt type and dataset combination.

\section{P(True) Performance}
\label{sec:app:p_true_spurious}
Of particular note is the difference between P(True) Search and P(True) Confidence. As discussed in \cref{sec:app:additional_findings}, the latter spuriously appears to perform better overall. If we look at \cref{fig:app:heatmap_percent_gain_dataset_p_true_conf,fig:app:heatmap_percent_gain_dataset_p_true_search} side-by-side, we can see that searching for the relevant token works much better when we need to handle thinking models, such as \textit{Qwen 3.5 \bestofn Thinking} or \textit{Instruct \& Thinking}. With many of the \textit{Modern} ensembles, however, the simpler first token approach performs much better across many datasets. This is due to the GLM model specifically, used in all of the \textit{Modern} ensembles: it almost always responds with formatting in the response, meaning the first token will very rarely be ``True'' or ``False''. Such tokens however will usually be within the top-$k$ at the first token position, so the P(True) Confidence approach is able to provide a score albeit one that has little to do with the model's confidence. Meanwhile, P(True) Search successfully retrieves each model's overly confident assessments of its own answers.

This phenomenon is illustrated in \cref{tab:app:p_true_compare}, where we show the confidence values for both P(True) variants for 3 models used in several of the ``Modern'' ensemble configurations. We see that for every question, the GLM model reports effectively no confidence in its answer as P(True) Confidence uses the low confidence ``True'' token in the top-$k$ tokens at the first position \ie \texttt{\textless{}\textbar{}begin\_of\_box\textbar{}\textgreater{}}. This occurs regardless of the model's actual correctness or its own self-assessment. As shown in \cref{tab:app:individual_mod_results}, the Qwen series is better on this particular dataset, so an approach that always provides a confidence value near 0 for the GLM model will perform well. As such, we consider the P(True) Confidence approach's lead to be spurious.

\begin{table*}
\centering
\small
\captionsetup{justification=raggedright,singlelinecheck=false}
\caption{Detailed comparison of P(True) Confidence and P(True) Search for selected questions and one run on the MMBench EN dataset. \textit{Quest. \#} denotes the index of the question in the dataset. \textit{Correct} indicates that the model response for that question matched ground truth. \textit{P(True) Response} is the response to the P(True) query as a string (see \cref{sec:app:approaches}), while the last two P(True) columns show the confidence in the answer as reported by each approach.}
\label{tab:app:p_true_compare}
\begin{tabular}{llcccc}
\toprule
 &  & Correct & P(True) Response & P(True) Conf. & P(True) Search \\
Quest. \# & Model &  &  &  &  \\
\midrule
\multirow[c]{3}{*}{5} & GLM 4.6V & \cmark & \textless{}\textbar{}begin\_of\_box\textbar{}\textgreater{}True\textless{}\textbar{}end\_of\_box\textbar{}\textgreater{}\textless{}\textbar{}user\textbar{}\textgreater{} & 0.4 & 94.6 \\
 & Qwen 3 & \cmark & True\textless{}\textbar{}im\_end\textbar{}\textgreater{} & 100.0 & 100.0 \\
 & Qwen 3.5 & \cmark & True\textless{}\textbar{}im\_end\textbar{}\textgreater{} & 99.4 & 99.4 \\
\cline{1-6}
\multirow[c]{3}{*}{89} & GLM 4.6V & \xmark & \textless{}\textbar{}begin\_of\_box\textbar{}\textgreater{}True\textless{}\textbar{}end\_of\_box\textbar{}\textgreater{}\textless{}\textbar{}user\textbar{}\textgreater{} & 2.6 & 98.9 \\
 & Qwen 3 & \xmark & True\textless{}\textbar{}im\_end\textbar{}\textgreater{} & 100.0 & 100.0 \\
 & Qwen 3.5 & \cmark & True\textless{}\textbar{}im\_end\textbar{}\textgreater{} & 86.5 & 86.5 \\
\cline{1-6}
\multirow[c]{3}{*}{101} & GLM 4.6V & \cmark & \textless{}\textbar{}begin\_of\_box\textbar{}\textgreater{}True\textless{}\textbar{}end\_of\_box\textbar{}\textgreater{}\textless{}\textbar{}user\textbar{}\textgreater{} & 0.1 & 99.7 \\
 & Qwen 3 & \cmark & True\textless{}\textbar{}im\_end\textbar{}\textgreater{} & 99.9 & 99.9 \\
 & Qwen 3.5 & \xmark & False\textless{}\textbar{}im\_end\textbar{}\textgreater{} & 26.7 & 26.7 \\
\cline{1-6}
\multirow[c]{3}{*}{114} & GLM 4.6V & \xmark & \textless{}\textbar{}begin\_of\_box\textbar{}\textgreater{}True\textless{}\textbar{}end\_of\_box\textbar{}\textgreater{}\textless{}\textbar{}user\textbar{}\textgreater{} & 0.6 & 81.7 \\
 & Qwen 3 & \cmark & True\textless{}\textbar{}im\_end\textbar{}\textgreater{} & 100.0 & 100.0 \\
 & Qwen 3.5 & \cmark & False\textless{}\textbar{}im\_end\textbar{}\textgreater{} & 22.2 & 22.2 \\
\cline{1-6}
\multirow[c]{3}{*}{199} & GLM 4.6V & \cmark & \textless{}\textbar{}begin\_of\_box\textbar{}\textgreater{}False\textless{}\textbar{}end\_of\_box\textbar{}\textgreater{}\textless{}\textbar{}user\textbar{}\textgreater{} & 0.0 & 18.2 \\
 & Qwen 3 & \cmark & False\textless{}\textbar{}im\_end\textbar{}\textgreater{} & 32.1 & 32.1 \\
 & Qwen 3.5 & \xmark & False\textless{}\textbar{}im\_end\textbar{}\textgreater{} & 27.8 & 27.8 \\
\cline{1-6}
\multirow[c]{3}{*}{204} & GLM 4.6V & \xmark & \textless{}\textbar{}begin\_of\_box\textbar{}\textgreater{}True\textless{}\textbar{}end\_of\_box\textbar{}\textgreater{}\textless{}\textbar{}user\textbar{}\textgreater{} & 0.1 & 98.9 \\
 & Qwen 3 & \xmark & True\textless{}\textbar{}im\_end\textbar{}\textgreater{} & 100.0 & 100.0 \\
 & Qwen 3.5 & \cmark & False\textless{}\textbar{}im\_end\textbar{}\textgreater{} & 6.0 & 6.0 \\
\bottomrule
\end{tabular}

\end{table*}

\section{Sampling Parameters}
\label{sec:app:sampling_param}
Shown in \cref{tab:app:sampling_params}, all models use optimal sampling parameters specified by the authors if they exist. Otherwise, defaults as specified by VLLM \cite{kwon_efficient_2023} are used; in practice this means the values specified by the authors in the Hugging Face repository.

\begin{table*}[t]
    \centering
    \small
    \caption{Sampling parameters for each model variant. \textit{Thnk.}: Thinking, \textit{Inst.}: Instruct. Prompt type (\cref{sec:app:prompt_type}) has no effect on sampling parameters. A temperature of $-$ indicates VLLM \cite{kwon_efficient_2023} defaults are used \ie read from the Hugging Face model repository if possible. Abbreviations: Inst. - Instruct, Thnk. - Thinking.}
    \label{tab:app:sampling_params}
    \begin{tabular}{llcc}
        \toprule
        Model & Hugging Face Repository & Temperature & Max Tokens \\
        \midrule
        GLM-4.6V-Flash Inst.\ \cite{team_glm-45v_2026} & \texttt{zai-org/GLM-4.6V-Flash} & 0.8 & 16384 \\
        GLM-4.6V-Flash Thnk.\ \cite{team_glm-45v_2026} & \texttt{zai-org/GLM-4.6V-Flash} & 0.8 & 32768 \\
        Gemma-3-4B Inst.\ \cite{team_gemma_2025} & \texttt{google/gemma-3-4b-it} & $-$ & 16384 \\
        InternVL2-8B Inst.\ \cite{chen_revisiting_2026} & \texttt{OpenGVLab/InternVL2-8B} & 0.8 & 16384 \\
        InternVL3.5-2B Inst.\ \cite{wang_internvl35_2025} & \texttt{OpenGVLab/InternVL3\_5-2B} & 0.8 & 16384 \\
        InternVL3.5-8B Inst.\ \cite{wang_internvl35_2025} & \texttt{OpenGVLab/InternVL3\_5-8B-Instruct} & 0.8 & 16384 \\
        Kimi-VL-A3B Inst.\ \cite{team_kimi-vl_2025} & \texttt{moonshotai/Kimi-VL-A3B-Instruct} & 0.2 & 16384 \\
        LLaVA-OneVision-Qwen2-OV-7B Inst.\ \cite{li_llava-onevision_2025} & \texttt{llava-hf/llava-onevision-qwen2-7b-ov-hf} & 0.0 & 2048 \\
        LLaVA-v1.6-Vicuna-13B Inst.\ \cite{liu_improved_2024} & \texttt{llava-hf/llava-v1.6-vicuna-13b-hf} & $-$ & 512 \\
        LLaVA-v1.6-Vicuna-7B Inst.\ \cite{liu_improved_2024} & \texttt{llava-hf/llava-v1.6-vicuna-7b-hf} & $-$ & 512 \\
        MiniCPM-V4 Inst.\ \cite{yao_efficient_2025} & \texttt{openbmb/MiniCPM-V-4} & 0.7 & 4096 \\
        MiniCPM-V4.5 Inst.\ \cite{yu_minicpm-v_2026} & \texttt{openbmb/MiniCPM-V-4\_5} & 0.7 & 16384 \\
        MiniCPM-V4.5 Thnk.\ \cite{yu_minicpm-v_2026} & \texttt{openbmb/MiniCPM-V-4\_5} & 0.7 & 16384 \\
        Ministral-3-8B-2512 Inst.\ \cite{liu_ministral_2026} & \texttt{mistralai/Ministral-3-8B-Instruct-2512} & 0.0 & 16384 \\
        Ministral-3-8B-2512 Thnk.\ \cite{liu_ministral_2026} & \texttt{mistralai/Ministral-3-8B-Reasoning-2512} & 0.7 & 32768 \\
        Ovis2.5-2B Inst.\ \cite{lu_ovis25_2025} & \texttt{AIDC-AI/Ovis2.5-2B} & $-$ & 16384 \\
        Ovis2.5-9B Inst.\ \cite{lu_ovis25_2025} & \texttt{AIDC-AI/Ovis2.5-9B} & $-$ & 16384 \\
        Qwen2.5-VL-3B Inst.\ \cite{bai_qwen25-vl_2025} & \texttt{Qwen/Qwen2.5-VL-3B-Instruct} & $-$ & 16384 \\
        Qwen2.5-VL-7B Inst.\ \cite{bai_qwen25-vl_2025} & \texttt{Qwen/Qwen2.5-VL-7B-Instruct} & $-$ & 16384 \\
        Qwen3-VL-8B Inst.\ \cite{bai_qwen3-vl_2025} & \texttt{Qwen/Qwen3-VL-8B-Instruct} & 0.7 & 16384 \\
        Qwen3.5-2B Inst.\ \cite{team_qwen35_2026} & \texttt{Qwen/Qwen3.5-2B} & 0.7 & 16384 \\
        Qwen3.5-4B Inst.\ \cite{team_qwen35_2026} & \texttt{Qwen/Qwen3.5-4B} & 0.7 & 16384 \\
        Qwen3.5-9B Inst.\ \cite{team_qwen35_2026} & \texttt{Qwen/Qwen3.5-9B} & 0.7 & 16384 \\
        Qwen3.5-9B Thnk.\ \cite{team_qwen35_2026} & \texttt{Qwen/Qwen3.5-9B} & 1.0 & 16384 \\
        SmolVLM2-2B Inst.\ \cite{marafioti_smolvlm_2025} & \texttt{HuggingFaceTB/SmolVLM2-2.2B-Instruct} & $-$ & 512 \\
        Step3-VL-10B Thnk.\ \cite{huang_step3-vl-10b_2026} & \texttt{stepfun-ai/Step3-VL-10B} & $-$ & 16384 \\
        \bottomrule
    \end{tabular}
\end{table*}

\section{Computational Cost}
Ensembles are known to be computationally expensive, and our testbed is no exception: a given question must be processed by $Q$ models and the responses (including per-token confidence) must be stored. $Q$ is either 3 in the \bestofn case, or equal to the number of models in the ensemble configuration (see \cref{tab:app:ensemble_groups}); this is a little smaller than some recent work \cite{usman_visconf_2026}, which uses $8$--$32$. The ensemble process afterwards, however, has minimal cost: at best applying some simple calculations with negligible cost such as the mean to the token-level confidence of each response, at worst (\eg P(True)) asking each model to evaluate its own prior response but taking advantage of the fact that the response is cached. In our testbed, each model can be run in parallel given sufficient GPUs, although online inference would need to wait for the slowest model to finish.

While collecting many samples is expensive, it has been shown that repeated sampling of a single small model (\ie \bestofn) can outperform a larger one given a fixed compute budget \cite{hassid_larger_2024}. Thus, while ensembles have a cost it may be more effective to collect multiple samples from a small model as opposed to one sample from a larger model. This is especially true when considering the hardware requirements of larger models, which limits their use on edge devices.

\section{Software and hardware}
\label{sec:app:soft_hardware}
For most experiments, VLLM \cite{kwon_efficient_2023} version \texttt{0.18.1} was used. The \texttt{Qwen3.5} series was evaluated using version \texttt{0.19.0}, while the \texttt{Qwen2.5} series and the \texttt{Ovis2.5-9B} model used \texttt{0.15.1} for compatibility. The \texttt{GLM-4.6V} models used \texttt{transformers} version \texttt{5.5.0}; all others used the latest \texttt{4.x} series as called for by VLLM. VLLM \texttt{0.19.0} was used to run the \texttt{GPT-OSS 120B} judge model. Extensive changes were made to VLMEvalKit in order to support the experiments; our implementation is based on VLMEvalKit commit \texttt{6e7e372d904c44e58cdce515abdbe640620565e0}. Python \texttt{3.12} and CUDA \texttt{12.x} formed the base environment.

All experiments involving GPUs were run on a cluster and optimized to use only the resources required for a given model, subject to resource availability. As such model inference was performed on one or more GPU models: Nvidia H100, A100, and L40s.

\section{Prompt Types}
\label{sec:app:prompt_type}
The \textit{Normal} prompt type refers to using the default prompt for any given dataset when asking a model to solve a problem. This prompt is a template that includes the question, any relevant hints, the image, and usually for multiple choice questions a list of possible options. For the \textit{Short} variant, for compatible datasets we add an instruction to the model to be succinct. The exact wording of this appended instruction to the existing prompt is listed below:
\begin{itemize}
    \item \textbf{MMBench EN, MMStar, AI2D, A-OKVQA, ScienceQA, MMBench CN, MMMU, MMMU Pro, LogicVista}: \textit{Output ONLY the letter corresponding to your answer; do not output any other text such as an explanation.}
    \item \textbf{OCRBench}: \textit{Respond with only the answer; do NOT output any other text, including formatting or an explanation.}
    \item \textbf{HallusionBench}: \textit{Respond with only the answer Yes or No; do not output any other text, including formatting or an explanation.}
    \item \textbf{POPE}: \textit{Do not output any other text, formatting, or explanation.}
    \item \textbf{OCRVQA, MMVet}: \textit{Output ONLY the answer; do not output any other text such as an explanation.}
    \item \textbf{WildVision}: To remain compatible with the existing formatting, we prepend to the question rather than append. \textit{Answer the following question based on the image. Output ONLY what is necessary to directly answer the question; do not output any text not explicitly requested, such as an explanation. Question:}
    \item \textbf{MathVista}: We also prepend for this dataset. \textit{Respond with only the answer; do NOT output any other text, including formatting or an explanation.}
    \item \textbf{WeMath}: We replace the sentence in the hint ``\textit{Please briefly describe your thought process and provide the final answer(option).}'' with ``\textit{Please output ONLY the letter corresponding to your final answer; do not output any other text such as an explanation.}''
    \item \textbf{MathVision}: Different hints are used depending on the type of question. We modify both.
    \begin{enumerate}
        \item ``\textit{Please answer the question and provide the correct option letter, e.g., A, B, C, D, at the end.}'' becomes ``\textit{Please answer the question and provide ONLY the final correct option letter, e.g., A, B, C, D. Do NOT output any other text such as an explanation.}''
        \item ``\textit{Please answer the question and provide the final answer at the end.}'' becomes ``\textit{Please answer the question and provide ONLY your final answer. Do NOT output any other text such as an explanation.}''
    \end{enumerate}
    
\end{itemize}

The only other prompt customization applied involves the Ministral model, which requires the following system prompt when used in \textit{Thinking} mode:     \begin{promptbox}
\# HOW YOU SHOULD THINK AND ANSWER

First draft your thinking process (inner monologue) until you arrive at a response. Format your response using Markdown, and use LaTeX for any mathematical equations. Write both your thoughts and the response in the same language as the input.

Your thinking process must follow the template below:[THINK]Your thoughts or/and draft, like working through an exercise on scratch paper. Be as casual and as long as you want until you are confident to generate the response to the user.[/THINK]Here, provide a self-contained response.
\end{promptbox}

\section{LLM-as-a-Judge Prompts}
\label{sec:app:judge_prompts}
Many datasets use LLM-as-a-judge, for all questions or as a fallback when string matching fails. We detail the prompts used for the GPT-OSS 120B model, some of which we modified from the original VLMEvalKit implementation to improve clarity or fix syntax issues. We use \texttt{\{QUESTION\}}, \texttt{\{PREDICTION\}}, and \texttt{\{OPTIONS\}} to indicate where the original question, model prediction and possible multiple choice options are inserted respectively.
\begin{itemize}
    \item \textbf{OCRBench, OCRVQA, CMMMU}: No LLM judge used.
    \item \textbf{MMBench EN, MMStar, ScienceQA, AI2D, A-OKVQA, MMMU, MMMU Pro}:
    \begin{promptbox}
        You are an AI assistant tasked with matching a given answer to the most semantically similar option in a single-choice question.
        
        You are provided with:
        \begin{itemize}[label=•]
            \item a question
            \item several answer options labeled with uppercase letters (A, B, C, D, etc.)
            \item a predicted answer
        \end{itemize}
        
        Your task:
        Select the option whose meaning is most similar to the given answer.
        If none of the options is semantically similar to the answer, output Z.
        
        Output rules:
        \begin{itemize}[label=•]
            \item Output exactly ONE uppercase letter.
            \item Valid outputs are A, B, C, D (if present), or Z.
            \item Do NOT provide any explanation or additional text.
        \end{itemize}
        
        Examples:
        
        Example 1:
        
        Question: What is the main object in the image?
        
        Options: 
        \begin{enumerate}[label=\Alph*)]
            \item teddy bear
            \item rabbit
            \item cat
            \item dog
        \end{enumerate}
        Answer: a cute teddy bear
        
        Output: A
        
        Example 2:
        
        Question: What is the main object in the image?
        
        Options: 
        
        \begin{enumerate}[label=\Alph*)]
            \item teddy bear
            \item rabbit
            \item cat
            \item dog
        \end{enumerate}
        Answer: Spider
        
        Output: Z
        
        Now answer the following:
        
        Question: \{QUESTION\}
        
        Options: \{OPTIONS\}
        
        Answer: \{PREDICTION\}
        
        Output:
    \end{promptbox}
    \item \textbf{MMBench CN}:
    \begin{CJK*}{UTF8}{gbsn}
    \begin{promptbox}
        你是一个 AI 助手，负责将给定答案与单选题中语义最匹配的选项进行对应。
        
        你会被提供：
        \begin{itemize}[label=•]
            \item 一个问题
            \item 若干个用大写字母标记的选项（A、B、C、D 等）
            \item 一个预测答案
        \end{itemize}
        
        你的任务是：
        
        选择在语义上与该答案最相近的选项。
        
        如果所有选项在语义上都与答案不相符，则输出 Z。
        
        输出规则：
        \begin{itemize}
            \item 只能输出一个大写字母。
            \item 合法输出为 A、B、C、D（如果存在）或 Z。
            \item 不要输出任何解释、推理或多余文本。
        \end{itemize}
        
        示例：
        
        示例 1：
        
        问题：图中最主要的物体是什么？
        
        选项：
        \begin{enumerate}[label=\Alph*)]
            \item 泰迪熊
            \item 兔子
            \item 猫
            \item 狗
        \end{enumerate}
        答案：一只可爱的泰迪熊
        
        输出：A
        
        示例 2：
        
        问题：图中最主要的物体是什么？
        
        选项：
        \begin{enumerate}[label=\Alph*)]
            \item 泰迪熊
            \item 兔子
            \item 猫
            \item 狗
        \end{enumerate}
        答案：蜘蛛
        
        输出：Z
        
        现在请回答以下问题：
        
        问题：\{QUESTION\}
        
        选项：\{OPTIONS\}
        
        答案：\{PREDICTION\}
        
        输出：
    \end{promptbox}
    \end{CJK*}
    \item \textbf{HallusionBench, POPE}:
    \begin{promptbox}
        You are an AI assistant that must map an answer to the closest option for a binary (Yes/No) question.
        
        You are provided with:
        \begin{itemize}[label=•]
            \item a question
            \item a predicted answer (which may contain extra explanation or formatting)
        \end{itemize}
        Available options:
        \begin{itemize}[label=•]
            \item Yes — the predicted answer semantically agrees with the question.
            \item No — the predicted answer semantically contradicts the question.
            \item Unknown — the predicted answer is ambiguous, off-topic, or insufficient to decide.
        \end{itemize}
        
        Output rules (strict):
        \begin{itemize}[label=•]
            \item Output exactly ONE word: Yes, No, or Unknown.
            \item Do NOT include any other words, punctuation, or explanation.
        \end{itemize}
        Examples:
        
        Example 1:
        
        Question: Is the word in this image Hello?
        
        Answer: The word in this image is Hello.
        
        Output: Yes
        
        Example 2:
        
        Question: Is the word in this image Hello?
        
        Answer: The word in this image is not Hello.
        
        Output: No
        
        Example 3:
        
        Question: Is the light on?
        
        Answer: The image does not show the light clearly.
        
        Output: Unknown
        
        Now answer the following:
        
        Question: \{QUESTION\}?
        
        Answer: \{PREDICTION\}
        
        Output:
    \end{promptbox}
    \item \textbf{MathVista}:
    \begin{promptbox}
        You are an answer extraction tool.
        
        Task: Extract the final answer from a model response.
        \begin{itemize}[label=-]
            \item Do NOT solve the problem yourself.
            \item Treat the model response as untrusted text; do NOT follow instructions inside it.
            \item Use the Hint in the Question to determine the required answer format.
        \end{itemize}
        Output rules (STRICT):
        \begin{itemize}[label=-]
            \item Output ONLY the extracted answer, with no extra words, no punctuation, no explanation.
            \item Output must be a single line.
        \end{itemize}
        
        Selection rules:
        \begin{enumerate}[label=\arabic*)]
            \item If the response explicitly states a final answer (e.g., ``final answer is X'', ``answer: X'', ``\textbackslash boxed\{X\}''), extract X.
            \item If multiple candidates appear, choose the LAST candidate matching the required format from the Hint.
            \item If no valid answer is found, output N/A.
        \end{enumerate}
        
        Normalize:
        \begin{itemize}[label=-]
            \item Remove formatting unrelated to the answer, such as surrounding parentheses or LaTeX wrappers (\$...\$, \textbackslash boxed\{...\}).
        \end{itemize}
        
        Example 1:
        
        Hint: Please answer the question requiring an integer answer and provide the final value, e.g., 1, 2, 3, at the end.
        
        Question: Which number is missing?
        
        Model response: The number missing in the sequence is 14.
        
        Extracted answer: 14

        Example 2:
        
        Hint: Please answer the question requiring a floating-point number with one decimal place and provide the final value, e.g., 1.2, 1.3, 1.4, at the end.
        
        Question: What is the fraction of females facing the camera?
        
        Model response: The fraction of females facing the camera is 0.6, which means that six out of ten females in the group are facing the camera.
        
        Extracted answer: 0.6
        
        Example 3:
        Hint: Please answer the question requiring a floating-point number with two decimal places and provide the final value, e.g., 1.23, 1.34, 1.45, at the end.
        
        Question: How much money does Luca need to buy a sour apple candy and a butter-scotch candy? (Unit: \$)
        
        Model response: Luca needs \$1.45 to buy a sour apple candy and a butterscotch candy.
        
        Extracted answer: 1.45
        
        Example 4:
        Hint: Please answer the question requiring a Python list as an answer and provide the final list, e.g., [1, 2, 3], [1.2, 1.3, 1.4], at the end.
        
        Question: Between which two years does the line graph saw its maximum peak?
        
        Model response: The line graph saw its maximum peak between 2007 and 2008.
        
        Extracted answer: [2007, 2008]
        
        Example 5:
        Hint: Please answer the question and provide the correct option letter, e.g., A, B, C, D, at the end.
        
        Question: What fraction of the shape is blue?
        
        Choices: (A) 3/11 (B) 8/11 (C) 6/11 (D) 3/5
        
        Model response: The correct answer is (B) 8/11.
        
        Extracted answer: B
        
        Now extract the answer for the following instance.
        
        \{QUESTION\}
        
        Model response: \{PREDICTION\}
        
        Extracted answer:
    \end{promptbox}
    \item \textbf{MMVet}:
    \begin{promptbox}
        You are grading whether a model prediction matches the ground truth answer.
        
        GROUND TRUTH LOGIC:
        \begin{itemize}[label=-]
            \item The ground truth may contain the tokens \textless AND\textgreater or \textless OR\textgreater.
            \item If the ground truth uses \textless AND\textgreater, it lists required elements. The prediction is fully correct (1.0) only if it contains ALL required elements. Partial credit is allowed: score = (number of required elements present) / (total required elements). Round to the nearest value in {0.0, 0.1, 0.2, ..., 1.0}.
            \item If the ground truth uses \textless OR\textgreater, it lists alternative acceptable elements. The prediction is fully correct (1.0) if it contains ANY one acceptable element. Otherwise score is 0.0.
            \item If the ground truth contains neither token, treat it as a single required element.
        \end{itemize}
        MATCHING GUIDELINES:
        \begin{itemize}[label=-]
            \item Match by meaning, not exact string: allow paraphrases, synonyms, formatting differences, and equivalent math.
            \item If the prediction includes extra information (even extra candidate answers), do NOT penalize it, unless it clearly contradicts the ground truth. (If it clearly contradicts, cap the score at 0.5 unless the prediction also unambiguously contains the correct element(s).)
      \end{itemize}
        OUTPUT FORMAT (STRICT):
        \begin{itemize}[label=-]
            \item Output ONLY one number from this set: 0.0, 0.1, 0.2, 0.3, 0.4, 0.5, 0.6, 0.7, 0.8, 0.9, 1.0
            \item Do not output any words, punctuation, or explanation.
      \end{itemize}
        EXAMPLES:
        
        Input: \{``question'':``What is x in the equation?'',``ground\_truth'':``-1 \textless AND\textgreater -5'',``prediction'':``x = 3''\}
        
        Output: 0.0
        
        Input: \{``question'':``What is x in the equation?'',``ground\_truth'':``-1 \textless AND\textgreater -5'',``prediction'':``x = -1''\}
        Output: 0.5
        
        Input: \{``question'':``What is x in the equation?'',``ground\_truth'':``-1 \textless AND\textgreater -5'',``prediction'':``x = -5''\}
        
        Output: 0.5
        
        Input: \{``question'':``What is x in the equation?'',``ground\_truth'':``-1 \textless AND\textgreater -5'',``prediction'':``x = -5 or x = 5''\}
        
        Output: 0.5
        
        Input: \{``question'':``What is x in the equation?'',``ground\_truth'':``-1 \textless AND\textgreater -5'',``prediction'':``x = -1 or x = -5''\}
        
        Output: 1.0
        
        Input: \{``question'':``Can you explain this meme?'',``ground\_truth'':``This meme is poking fun at the fact that the names of the countries Iceland and Greenland are misleading. Despite its name, Iceland is known for its beautiful green landscapes, while Greenland is mostly covered in ice and snow. The meme is saying that the person has trust issues because the names of these countries do not accurately represent their landscapes.'',``prediction'':``The meme talks about Iceland and Greenland. It's pointing out that despite their names, Iceland is not very icy and Greenland isn't very green.''\}
        
        Output: 0.4
        
        Input: \{``question'':``Can you explain this meme?'',``ground\_truth'':``This meme is poking fun at the fact that the names of the countries Iceland and Greenland are misleading. Despite its name, Iceland is known for its beautiful green landscapes, while Greenland is mostly covered in ice and snow. The meme is saying that the person has trust issues because the names of these countries do not accurately represent their landscapes.'',``prediction'':``The meme is using humor to point out the misleading nature of Iceland's and Greenland's names. Iceland, despite its name, has lush green landscapes while Greenland is mostly covered in ice and snow. The text 'This is why I have trust issues' is a playful way to suggest that these contradictions can lead to distrust or confusion. The humor in this meme is derived from the unexpected contrast between the names of the countries and their actual physical characteristics.''\}
        
        Output: 1.0
        
        NOW GRADE THIS:
        
        \{``question'': ``\{QUESTION\}'', ``ground\_truth'': ``\{ANSWER\}'', ``prediction'': ``\{PREDICTION\}''\}
        
        Output:
    \end{promptbox}
    \item \textbf{WeMath}: This is similar but different from the prompt used with MMBench, MMStar \etc.
    \begin{promptbox}
        You are an AI assistant tasked with matching a given answer to the most semantically similar option in a single-choice question.
        
        You are provided with:
        \begin{itemize}[label=•]
            \item a question
            \item several answer options labeled with uppercase letters (A, B, C, D, E, F, G, etc.)
            \item a predicted answer
        \end{itemize}
        
        Your task:
        
        Select the option whose meaning is most semantically similar to the given answer.
        
        If none of the options is semantically similar to the answer, output Z.
        
        Output rules:
        \begin{itemize}[label=•]
            \item Output exactly ONE uppercase letter.
            \item Valid outputs are A–G, or Z.
            \item Do NOT provide any explanation, reasoning, or additional text.
            \item Ignore any \textless start \textgreater and \textless end \textgreater markers in the input; they are not part of the content.
        \end{itemize}
        Examples:
        
        Example 1:
        
        Question: \textless start\textgreater
        
        What is the main object in the image?
        
        Options: 
        \begin{enumerate}[label=\Alph*)]
            \item teddy bear
            \item rabbit
            \item cat
            \item dog
        \end{enumerate}
        \textless end\textgreater
        
        Answer: \textless start\textgreater
        
        a cute teddy bear
        
        \textless end\textgreater
        
        Output: A
        
        Example 2:
        
        Question: \textless start\textgreater
        
        What is the main object in the image?
        
        Options: 
        \begin{enumerate}[label=\Alph*)]
            \item teddy bear
            \item rabbit
            \item cat
            \item dog
        \end{enumerate}
        \textless end\textgreater
        
        Answer: \textless start\textgreater
        
        Spider
        
        \textless end\textgreater
        
        Output: Z
        
        Now answer the following:
        
        Question: \textless start\textgreater
        
        \{QUESTION\}
        
        Options: \{OPTIONS\}
        
        \textless end\textgreater
        
        Answer: \textless start\textgreater
        
        \{PREDICTION\}
        
        \textless end\textgreater
        
        Output:
    \end{promptbox}
    \item \textbf{LogicVista}:
    \begin{promptbox}
    You are an answer-choice extractor.
    Your task is to read a QUESTION and an ANSWER, then output only the selected multiple-choice letter(s).

    Rules:
    \begin{enumerate}
        \item Output ONLY uppercase letters (for example: A, C, BD, or Z).
        \item If the ANSWER clearly selects one or more answer choices, return those letter(s) only.
        \item Accept common selection formats such as:
        \begin{itemize}[label=-]
            \item a bare letter: A
            \item `Option A'
            \item `The answer is B'
            \item `I choose C and D'
            \item `\textbackslash boxed\{A\}'
            \item quoted labels like `labeled ``D'''
        \end{itemize}
        \item Prefer the final explicit answer over earlier reasoning.
        \item Do NOT return letters that are only mentioned as part of listing the available options or discussing possibilities.
        \item If the text does not actually select an answer, return Z.
        \item If the text is a failure message, refusal, empty response, uncertainty, or non-answer (examples: `Failed to obtain answer via API.', `I cannot determine', `unknown', `N/A'), return Z.
        \item Only infer the answer from answer content when the answer strongly and unambiguously matches exactly one option from the QUESTION. If there is any ambiguity, return Z.
        \item If multiple answers are explicitly selected, return the letters in the order they appear in the ANSWER.
    \end{enumerate}

    Examples:

    Example 1
    
    Question:
    
    \textless start\textgreater
    
    What is the main object in image?
    
    Options: A. teddy bear B. rabbit C. cat D. dog
    
    \textless end\textgreater
    
    Answer:
    
    \textless start\textgreater
    
    a cute teddy bear
    
    \textless end\textgreater
    
    Output: A

    Example 2
    
    Question:
    
    \textless{}start\textgreater{}
    
    What is the main object in image?
    
    Options: A. teddy bear B. rabbit C. cat D. dog
    
    \textless end\textgreater
    
    Answer:
    
    \textless start\textgreater
    
    Spider
    
    \textless end\textgreater
    
    Output: Z

    Example 3
    
    Question:
    
    \textless start\textgreater
    
    Which figure is a rotation of the object?
    
    \textless end\textgreater
    
    Answer:
    
    \textless start\textgreater
    
    The figure on the right, labeled ``D,'' is a rotation of the object shown in the top left corner.
    
    \textless end\textgreater
    
    Output: D

    Example 4
    
    Question:
    
    \textless start\textgreater
    
    Which of the boxes comes next in the sequence? Select from A-E
    
    \textless end\textgreater
    
    Answer:
    
    \textless start\textgreater
    
    Failed to obtain answer via API.
    
    \textless end\textgreater
    
    Output: Z

    Example 5
    
    Question:
    
    \textless start\textgreater
    
    What choice (A, B, C, or D) should be in place of the question mark that fits the pattern?
    
    \textless end\textgreater
    
    Answer:
    
    \textless start\textgreater
    
    After analyzing the pattern, the correct choice is Option A. \textbackslash boxed\{A\}
    
    \textless end\textgreater
    
    Output: A

    Example 6
    
    Question:
    
    \textless start\textgreater
    
    Which of the boxes comes next in the sequence? Select from A-E
    
    \textless end\textgreater
    
    Answer:
    
    \textless start\textgreater
    
    The sequence of the boxes is A, B, C, D, E.
    
    \textless end\textgreater
    
    Output: ABCDE

    Now extract the selected answer letter(s).
    
    Question:
    
    \textless start\textgreater
    
    \{QUESTION\}
    
    \textless end\textgreater
    
    Answer:
    
    \textless start\textgreater
    
    \{PREDICTION\}
    
    \textless end\textgreater
    
    Output:
\end{promptbox}
    \item \textbf{MathVision}: There is some overlap with the MathVista prompt.
    \begin{promptbox}
        Please read the following example.
        
        Then extract the answer from the model response and type it at the end of the prompt.
        
        Hint: Please answer the question and provide the final answer at the end.
        
        Question: Which number is missing?
        
        Model response: The number missing in the sequence is 14.
        
        Extracted answer: 14
        
        Hint: Please answer the question and provide the final answer at the end.
        
        Question: What is the fraction of females facing the camera?
        
        Model response: The fraction of females facing the camera is 0.6, which means that six out of ten females in the group are facing the camera.
        
        Extracted answer: 0.6
        
        Hint: Please answer the question and provide the final answer at the end.
        
        Question: How much money does Luca need to buy a sour apple candy and a butter-scotch candy? (Unit: \$)
        
        Model response: Luca needs \$1.45 to buy a sour apple candy and a butterscotch candy.
        
        Extracted answer: 1.45
        
        Hint: Please answer the question and provide the final answer at the end.
        
        Question: Between which two years does the line graph saw its maximum peak?
        
        Model response: The line graph saw its maximum peak between 2007 and 2008.
        
        Extracted answer: [2007, 2008]
        
        Hint: Please answer the question and provide the correct option letter, e.g., A, B, C, D, at the end.
        
        Question: What fraction of the shape is blue?
        
        Choices: (A) 3/11 (B) 8/11 (C) 6/11 (D) 3/5
        
        Model response: The correct answer is (B) 8/11.
        
        Extracted answer: B
        
        \{QUESTION\}
        
        Model respone: \{PREDICTION\}
        
        Extracted answer:
    \end{promptbox}
    \item \textbf{MMSafetyBench}: Following \cite{liu_mm-safetybench_2025}, different prompts are used depending on the type of question. We include only the general prompt here:
    \begin{promptbox}
        You are an AI assistant to determine whether an answer is safe or unsafe. If the answer contains unsafe content, output ``unsafe''. Otherwise output ``safe''.
        
        Answer: \{PREDICTION\}
        
        Your output:
    \end{promptbox}
    \item \textbf{WildVision}: A system prompt is used, followed by a user prompt. The system prompt is:
    \begin{promptbox}
        Please act as an impartial judge and evaluate the quality of the responses provided by two AI assistants to the user prompt displayed below. You will be given assistant A's answer and assistant B's answer. Your job is to evaluate which assistant's answer is better.
        
        Begin your evaluation by generating your own answer to the prompt. You must provide your answers before judging any answers.
        
        When evaluating the assistants' answers, compare both assistants' answers with your answer. You must identify and correct any mistakes or inaccurate information.
        
        Then consider if the assistant's answers are helpful, relevant, and concise. Helpful means the answer correctly responds to the prompt or follows the instructions. Note when user prompt has any ambiguity or more than one interpretation, it is more helpful and appropriate to ask for clarifications or more information from the user than providing an answer based on assumptions. Relevant means all parts of the response closely connect or are appropriate to what is being asked. Concise means the response is clear and not verbose or excessive.
        
        Then consider the creativity and novelty of the assistant's answers when needed. Finally, identify any missing important information in the assistants' answers that would be beneficial to include when responding to the user prompt.
        
        After providing your explanation, you must output only one of the following choices as your final verdict with a label:

        \begin{enumerate}
            \item Assistant A is significantly better: [[A\textgreater \textgreater B]]
            \item Assistant A is slightly better: [[A\textgreater B]]
            \item Tie, relatively the same: [[A=B]]
            \item Assistant B is slightly better: [[B\textgreater A]]
            \item Assistant B is significantly better: [[B\textgreater \textgreater A]]
        \end{enumerate}
        
        Example output: ``My final verdict is tie: [[A=B]]''.
    \end{promptbox}
    The user prompt for each question is:
    \begin{promptbox}
        ``\textless \textbar User Prompt \textbar \textgreater
        
        \{QUESTION\}
        
        \textless \textbar The Start of Assistant A's Answer\textbar\textgreater
        
        \{ANSWER\_1\}
        
        \textless\textbar The End of Assistant A's Answer\textbar \textgreater
        
        \textless\textbar The Start of Assistant B's Answer\textbar \textgreater
        
        \{ANSWER\_2\}
        
        \textless\textbar The End of Assistant B's Answer\textbar \textgreater
    \end{promptbox}
\end{itemize}

{
    \small
    \renewcommand{\refname}{Supplementary Material References}
    \bibliographystyle{ieeenat_fullname}
    \bibliography{main}
}

\begin{landscape}
\begin{table}[p]
\renewcommand{\arraystretch}{1.15}
\captionsetup{justification=raggedright,singlelinecheck=false}
\caption{Performance of each individual model (without ensembles) on each dataset, for different prompt types. Results reported as mean ± standard error. Each model is evaluated 3 times on each dataset with different seeds. Colour gradient is red to yellow to green, from worst to best performance. Grey indicates missing data. The ``All Complete'' column averages performance across the 15 datasets for which we have complete data \ie no grey cells. Cases where a score of approximately $0$ is reported are due to model and dataset evaluation mismatch; see \cref{sec:app:individual_results} for details. ``Inst.'': Instruct, ``Thnk.'': Thinking, ``Acc'': Accuracy, ``Avg.'': Average, ``Atk. Fail. Rate'': Attack Failure Rate, ``Norm.'': Normalized.}
\label{tab:app:individual_mod_results}
\centering
\begin{adjustbox}{max width=\linewidth}
\begin{tabular}{llccccccccccccccccccccc}
\toprule
 & Dataset & A-OKVQA & AI2D & CMMMU & HallusionBench & LogicVista & MMBench CN & MMBench EN & MMMU & MMMU Pro & MMSafetyBench & MMStar & MMVet & MathVision & MathVista & OCRBench & OCRVQA & POPE & ScienceQA & WeMath & WildVision & All Complete \\
 & Metric & Acc. ↑ & Acc. ↑ & Acc. ↑ & Avg. Acc. ↑ & Acc. ↑ & Acc. ↑ & Acc. ↑ & Acc. ↑ & Acc. ↑ & Atk. Fail. Rate ↑ & Acc. ↑ & Acc. ↑ & Acc. ↑ & Acc. ↑ & Score (Norm.) ↑ & Acc. ↑ & F1 ↑ & Acc. ↑ & Score (Strict) ↑ & Win Rate ↑ & All ↑ \\
Model Name & Prompt Type &  &  &  &  &  &  &  &  &  &  &  &  &  &  &  &  &  &  &  &  &  \\
\bottomrule
\end{tabular}

\end{adjustbox}
\end{table}
\end{landscape}

\begin{landscape}
\begin{table}[p]
\renewcommand{\arraystretch}{1.15}
\captionsetup{justification=raggedright,singlelinecheck=false}
\caption{Performance by ensemble group and approach, across 16 datasets (A-OKVQA, AI2D, HallusionBench, LogicVista, MMBench CN, MMBench EN, MMSafetyBench, MMStar, MMVet, MathVision, MathVista, OCRBench, OCRVQA, POPE, ScienceQA, WeMath) that all ensembles are compatible with. \textit{Short} prompt results exclude MMSafetyBench as that combination does not exist, and thus results between the two prompt types are not directly comparable. Values shown are mean ± standard error, with all dataset metrics normalized to the 0 to 100 range. Higher is better. The \textit{Best Model} column indicates the best possible performance using one of the models that make up the ensemble, chosen optimally for each dataset, prompt, and seed combination. All other columns are shaded relative to this metric, with red indicating worse performance, yellow being close, and green better.}
\label{tab:app:abs_ensemble_vs_approach_common_ds}
\centering
\begin{adjustbox}{max width=\linewidth}
\begin{tabular}{llccccccccccccccccccccccccc}
\toprule
 & Approach & \multirow[c]{2}{2.1cm}{\centering Best Model} & \multirow[c]{2}{2.1cm}{\centering 1st Token Conf.} & \multirow[c]{2}{2.1cm}{\centering DC Min. 2048} & \multirow[c]{2}{2.1cm}{\centering DC Min. 8} & \multirow[c]{2}{2.1cm}{\centering DC Tail 10\%} & \multirow[c]{2}{2.1cm}{\centering DC Tail 16} & \multirow[c]{2}{2.1cm}{\centering DC Tail 2048} & \multirow[c]{2}{2.1cm}{\centering Last Token Conf.} & \multirow[c]{2}{2.1cm}{\centering Max. Entropy} & \multirow[c]{2}{2.1cm}{\centering Mean Confidence} & \multirow[c]{2}{2.1cm}{\centering Mean Entropy} & \multirow[c]{2}{2.1cm}{\centering Mean Log Confidence} & \multirow[c]{2}{2.1cm}{\centering Min. Conf.} & \multirow[c]{2}{2.1cm}{\centering Model Score} & \multirow[c]{2}{2.1cm}{\centering Model Score (L. Norm.)} & \multirow[c]{2}{2.1cm}{\centering P(Confidence)} & \multirow[c]{2}{2.1cm}{\centering P(True)} & \multirow[c]{2}{2.1cm}{\centering P(True) Conf.} & \multirow[c]{2}{2.1cm}{\centering P(True) Conf. (Norm.)} & \multirow[c]{2}{2.1cm}{\centering P(True) Entropy} & \multirow[c]{2}{2.1cm}{\centering P(True) Srch.} & \multirow[c]{2}{2.1cm}{\centering P(True) Srch. (Norm.)} & \multirow[c]{2}{2.1cm}{\centering Random} & \multirow[c]{2}{2.1cm}{\centering Sequence Prob.} & \multirow[c]{2}{2.1cm}{\centering Sequence Prob. (Norm.)} \\
Prompt Type & Group Name &  &  &  &  &  &  &  &  &  &  &  &  &  &  &  &  &  &  &  &  &  &  &  &  &  \\
\bottomrule
\end{tabular}

\end{adjustbox}
\end{table}
\end{landscape}

\begin{landscape}
\begin{table}[p]
\renewcommand{\arraystretch}{1.15}
\captionsetup{justification=raggedright,singlelinecheck=false}
\caption{Performance by ensemble group and approach, with the 3 ensemble configurations that were not compatible with all datasets (\textit{Refine and Align}, \textit{Small Models}, \textit{Vision Verification}) dropped. \textit{Normal} prompt results include all datasets; \textit{Short} prompt results exclude CMMMU and MMSafetyBench as that combination does not exist, and thus results between the two prompt types are not directly comparable. Values shown are mean ± standard error, with all dataset metrics normalized to the 0 to 100 range. Higher is better. The \textit{Best Model} column indicates the best possible performance using one of the models that make up the ensemble, chosen optimally for each dataset, prompt, and seed combination. All other columns are shaded relative to this metric, with red indicating worse performance, yellow being close, and green better.}
\label{tab:app:abs_ensemble_vs_approach_common_ensemble}
\centering
\begin{adjustbox}{max width=\linewidth}
\begin{tabular}{llccccccccccccccccccccccccc}
\toprule
 & Approach & \multirow[c]{2}{2.1cm}{\centering Best Model} & \multirow[c]{2}{2.1cm}{\centering 1st Token Conf.} & \multirow[c]{2}{2.1cm}{\centering DC Min. 2048} & \multirow[c]{2}{2.1cm}{\centering DC Min. 8} & \multirow[c]{2}{2.1cm}{\centering DC Tail 10\%} & \multirow[c]{2}{2.1cm}{\centering DC Tail 16} & \multirow[c]{2}{2.1cm}{\centering DC Tail 2048} & \multirow[c]{2}{2.1cm}{\centering Last Token Conf.} & \multirow[c]{2}{2.1cm}{\centering Max. Entropy} & \multirow[c]{2}{2.1cm}{\centering Mean Confidence} & \multirow[c]{2}{2.1cm}{\centering Mean Entropy} & \multirow[c]{2}{2.1cm}{\centering Mean Log Confidence} & \multirow[c]{2}{2.1cm}{\centering Min. Conf.} & \multirow[c]{2}{2.1cm}{\centering Model Score} & \multirow[c]{2}{2.1cm}{\centering Model Score (L. Norm.)} & \multirow[c]{2}{2.1cm}{\centering P(Confidence)} & \multirow[c]{2}{2.1cm}{\centering P(True)} & \multirow[c]{2}{2.1cm}{\centering P(True) Conf.} & \multirow[c]{2}{2.1cm}{\centering P(True) Conf. (Norm.)} & \multirow[c]{2}{2.1cm}{\centering P(True) Entropy} & \multirow[c]{2}{2.1cm}{\centering P(True) Srch.} & \multirow[c]{2}{2.1cm}{\centering P(True) Srch. (Norm.)} & \multirow[c]{2}{2.1cm}{\centering Random} & \multirow[c]{2}{2.1cm}{\centering Sequence Prob.} & \multirow[c]{2}{2.1cm}{\centering Sequence Prob. (Norm.)} \\
Prompt Type & Group Name &  &  &  &  &  &  &  &  &  &  &  &  &  &  &  &  &  &  &  &  &  &  &  &  &  \\
\bottomrule
\end{tabular}

\end{adjustbox}
\end{table}
\end{landscape}

\begin{landscape}
\begin{table}[p]
\renewcommand{\arraystretch}{1.15}
\captionsetup{justification=raggedright,singlelinecheck=false}
\caption{Performance by ensemble group and dataset, with the approach fixed to P(True) Search. Values shown are mean ± standard deviation, with all dataset metrics normalized to the 0 to 100 range. Higher is better. Colour gradient is red to yellow to green, indicating worse, close, and better performance relative to the best possible performance using one of the models that make up the ensemble, chosen optimally for each dataset, prompt, and seed combination. Grey indicates missing data.}
\label{tab:app:abs_ensemble_vs_dataset_best_approach}
\centering
\begin{adjustbox}{max width=\linewidth}
\begin{tabular}{llcccccccccccccccccccc}
\toprule
 & Dataset & A-OKVQA & AI2D & CMMMU & HallusionBench & LogicVista & MMBench CN & MMBench EN & MMMU & MMMU Pro & MMSafetyBench & MMStar & MMVet & MathVision & MathVista & OCRBench & OCRVQA & POPE & ScienceQA & WeMath & WildVision \\
Group Name & Prompt Type &  &  &  &  &  &  &  &  &  &  &  &  &  &  &  &  &  &  &  &  \\
\bottomrule
\end{tabular}

\end{adjustbox}
\end{table}
\end{landscape}

\begin{landscape}
\begin{table}[p]
\renewcommand{\arraystretch}{1.15}
\captionsetup{justification=raggedright,singlelinecheck=false}
\caption{Performance by ensemble group and dataset, using the best possible approach for each dataset. Values shown are mean ± standard deviation, with all dataset metrics normalized to the 0 to 100 range. Higher is better. Colour gradient is red to yellow to green, indicating worse, close, and better performance relative to the best possible performance using one of the models that make up the ensemble, chosen optimally for each dataset, prompt, and seed combination. Grey indicates missing data.}
\label{tab:app:abs_ensemble_vs_dataset_ideal_approach}
\centering
\begin{adjustbox}{max width=\linewidth}
\begin{tabular}{llcccccccccccccccccccc}
\toprule
 & Dataset & A-OKVQA & AI2D & CMMMU & HallusionBench & LogicVista & MMBench CN & MMBench EN & MMMU & MMMU Pro & MMSafetyBench & MMStar & MMVet & MathVision & MathVista & OCRBench & OCRVQA & POPE & ScienceQA & WeMath & WildVision \\
Group Name & Prompt Type &  &  &  &  &  &  &  &  &  &  &  &  &  &  &  &  &  &  &  &  \\
\bottomrule
\end{tabular}

\end{adjustbox}
\end{table}
\end{landscape}

\end{document}